\documentclass[10pt,twocolumn,letterpaper]{article}

\usepackage{iccv}
\usepackage{times}
\usepackage{epsfig}
\usepackage{graphicx}
\usepackage{amsmath}
\usepackage{amssymb}
\usepackage{tabularx}
\usepackage{booktabs}

\usepackage{cite}
\usepackage{amsfonts}
\usepackage{algorithmic}
\usepackage{textcomp}
\usepackage{makecell}
\usepackage{multirow}
\usepackage{subcaption}
\usepackage{amsthm}
\usepackage[table]{xcolor}

\definecolor{dino}{RGB}{249,231,227}
\definecolor{dino_text}{RGB}{209,154,128}

\usepackage{setspace}
\usepackage[ruled,norelsize]{algorithm2e}

\definecolor{commentcolor}{RGB}{110,154,155}   
\newcommand{\PyComment}[1]{\ttfamily\textcolor{commentcolor}{\# #1}}  
\newcommand{\PyCode}[1]{\ttfamily\textcolor{black}{#1}} 

\usepackage[pagebackref=true,breaklinks=true,letterpaper=true,colorlinks,bookmarks=false]{hyperref}

\iccvfinalcopy 

\def\iccvPaperID{6302} 

\begin{document}

\title{Multiple Instance Learning Framework with Masked Hard Instance Mining for Whole Slide Image Classification}

\author{Wenhao Tang$^1$ ~~ Sheng Huang$^1$$\thanks{Corresponding Author.}$ ~~ Xiaoxian Zhang$^1$ ~~ Fengtao Zhou$^2$ ~~ Yi Zhang$^1$ ~~ Bo Liu$^3$\\
{ $^1$ Chongqing University ~~ $^2$ The Hong Kong University of Science and Technology} \\ {$^3$ Walmart Global Tech}\\
{\tt\small \{whtang, huangsheng, zhangxiaoxian, zhangyii\}@cqu.edu.cn}\\ 
{\tt\small fzhouaf@connect.ust.hk, kfliubo@gmail.com}
}

\maketitle
\ificcvfinal\thispagestyle{empty}\fi

\begin{abstract}
   The whole slide image (WSI) classification is often formulated as a multiple instance learning (MIL) problem. 
    Since the positive tissue is only a small fraction of the gigapixel WSI,
    existing MIL methods intuitively focus on identifying salient instances via attention mechanisms.
    However, this leads to a bias towards easy-to-classify instances while neglecting hard-to-classify instances. 
    Some literature has revealed that hard examples are beneficial for modeling a discriminative boundary accurately. 
    By applying such an idea at the instance level, 
    we elaborate a novel MIL framework with masked hard instance mining (MHIM-MIL),
    which uses a Siamese structure (Teacher-Student) with a consistency constraint to explore the potential hard instances.
    With several instance masking strategies based on attention scores, MHIM-MIL employs a momentum teacher to implicitly mine hard instances for training the student model, which can be any attention-based MIL model.
    This counter-intuitive strategy essentially enables the student to learn a better discriminating boundary.
    Moreover, the student is used to update the teacher with an exponential moving average (EMA), which in turn identifies new hard instances for subsequent training iterations and stabilizes the optimization. 
    Experimental results on the CAMELYON-16 and TCGA Lung Cancer datasets demonstrate that MHIM-MIL outperforms other latest methods in terms of performance and training cost.
    The code is available at:~\href{https://github.com/DearCaat/MHIM-MIL}{https://github.com/DearCaat/MHIM-MIL}.
\end{abstract}

\section{Introduction}
\label{sec:intro}
Histopathological image analysis plays a crucial role in modern medicine, particularly in the treatment of cancer, where it serves as the gold standard for diagnosis~\cite{pinckaers2020intro_pami,li2021dt,zhao2022setmil,lu2021nature}. 
Digitalizating pathological images into Whole Slide Images (WSIs) through digital slide scanner has opened new avenues for computer-aided analysis~\cite{shao2021transmil,chikontwe2021dual}.
Due to the huge size of a WSI and the lack of pixel-level annotations, histopathological image analysis is commonly formulated as a multiple instance learning (MIL) task~\cite{srinidhi2021survey,maron1997mil_1,dietterich1997mil_2}.
In MIL, each WSI (or slide) is a bag containing thousands of unlabeled instances (patches) cropped from the slide. With at least one instance being disease positive, the bag is deemed positive, otherwise negative. 

\begin{figure}[t]
\centering
\includegraphics[width=\linewidth]{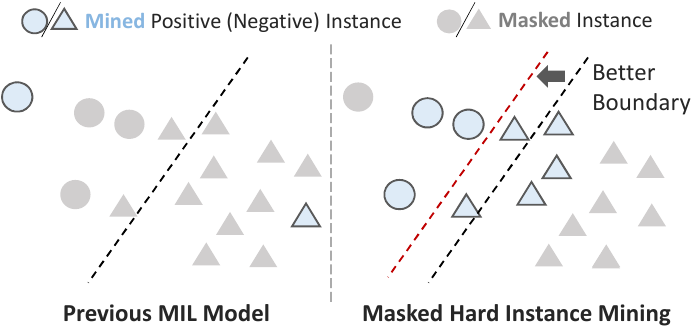}
\caption{\textbf{Left:} Previous MIL models focus on the more salient instances. \textbf{Right:} MHIM-MIL mines an amount of hard-to-classify instances to learn a better boundary.}
\label{fig:intro}
\end{figure}
However, the number of slides is limited and each slide contains a mass of instances with a low positive proportion. This imbalance would hinder the inference of bag labels~\cite{zhang2022dtfd,ilse2018attention}.
To alleviate this issue, several WSI classification methods~\cite{ilse2018attention,shao2021transmil,li2021dual,li2021dt,chen2022hipt} employ an attention mechanism to aggregate salient instance features into a bag-level feature for WSI classification.
Furthermore, some MIL frameworks~\cite{zhang2022dtfd,li2021dual,clam,xu2019camel} focus on the more salient instances in the bag and leverage them to facilitate WSI classification.
For instance, existing frameworks~\cite{zhang2022dtfd,xu2019camel} propose to only select the instances that correspond to the top $K$ highest or lowest attention scores~\cite{xu2019camel,li2021dual} or patch probabilities~\cite{zhang2022dtfd} for yielding high-quality bag embedding for both training and testing.

These salient instances are actually ``easy-to-classify" instances, which are not optimal for training a discriminative WSI classification model.
In conventional machine learning, such as Support Vector Machines (SVM)~\cite{hearst1998svm}, samples near the category distribution boundary are more challenging to classify, but are more useful for depicting the classification boundary, as illustrated in Figure~\ref{fig:intro}.  
Moreover, other deep learning works~\cite{sun2019him_3,tan2022him_4,sheng2020hsm,schroff2015facenet} also reveal that mining hard samples for training can improve the generalization abilities of models.
By applying such an idea at the instance level, we can better highlight the ``hard-to-classify" instances that facilitate MIL model training, and benefit the final WSI classification.
However, the lack of instance labels poses a challenge to the direct application of traditional hard sample mining strategies at the instance level.

To address this issue, 
we present a novel MIL framework based on masked hard instance mining strategies (MHIM) named MHIM-MIL. 
The main idea of MHIM is to mask out the instances with high attention scores to highlight the hard instances for model training. 
Based on this, we incorporate two other instance masking strategies to enhance training efficiency and mitigate the over-fitting risk.
Another key design of MHIM-MIL is an instance attention generator based on a Siamese structure (Teacher-Student)~\cite{bromley1993siamese,chen2021simsiam}.
In MHIM-MIL, the MIL-based WSI classification model is the student network, which aggregates hard instances mined by a momentum teacher with different instance masking strategies. The momentum teacher is updated using an exponential moving average (EMA) of the student model. 
Moreover, the framework is optimized by inducing a consistency constraint that explores more supervised information beyond the limited slide label. 
Unlike the conventional MIL frameworks~\cite{zhang2022dtfd,xu2019camel}, which adopt complex cascade gradient-updating structures, our method is more simple and does not require additional parameters. It not only improves efficiency but also provides improved performance stability.
The contribution of this paper is summarized as follows,
\begin{itemize}
    \item We propose a simple and efficient MIL framework with masked hard instance mining named MHIM-MIL. It implicitly mines hard instances with instance attention for training a more discriminative MIL model. Extensive experiments on two WSI datasets validate that MHIM boosts different MIL models and outperforms other latest methods in terms of performance and training cost.
    \item We propose several hybrid instance masking strategies for indirectly mining hard instances in MIL. These strategies not only address the reliance problem of conventional methods on instance-level supervision but also enhance the training efficiency of the model and mitigate the over-fitting risk. 
    \item With the Siamese structure, we introduce a parameter-free momentum teacher to obtain instance attention scores more efficiently and stably. 
    Moreover, we employ a consistency-based iterative optimization to improve the discriminability of both models progressively.

\end{itemize}


\section{Related Work}
\subsection{Multiple Instance Learning in WSI Analysis}
Multiple Instance Learning (MIL)~\cite{dietterich1997mil_2} has been widely used in WSI analysis with its unique learning paradigm in recent years~\cite{manivannan2017subcategory,xu2014weakly,xu2019camel,li2021dual,shao2021transmil,tong2014multiple}. 
MIL is a weakly supervised learning framework that utilizes coarse-grained bag labels for training instead of fine-grained instance annotations.
Previous algorithms can be broadly categorized into two groups: instance-level~\cite{campanella2019clinical,hou2016patch,xu2019camel,instance_mil_1} and embedding-level~\cite{chikontwe2021dual,wu2021combining,zhang2022dtfd,sharma2021cluster,wang2018revisiting}. 
The former obtain instance labels and aggregate them to obtain the bag label, whereas the latter aggregate all instance features into a high-level bag embedding for bag prediction.
Most embedding-level methods share the basic idea of AB-MIL~\cite{ilse2018attention}, which employs learnable weights to aggregate salient instance features into bag embedding.
Furthermore, some MIL frameworks~\cite{zhang2022dtfd,li2021dual,clam,xu2019camel} mine more salient instances making classification easier and facilitating classification. 
For example, Lu \etal selected the most salient instances based on their attention scores (e.g., maximum and minimum scores) to compute instance-level loss and improve performance~\cite{clam}.
Zhang \etal proposed a class activation map (CAM) based on the AB-MIL paradigm to better mine salient instances and used AB-MIL to aggregate them into bag embedding~\cite{zhang2022dtfd}. 
In addition, feature clustering methods~\cite{Zhang_2022_BMVC,sharma2021cluster,wang2019weakly} computed cluster centroids of all feature embeddings and used representative embeddings for the final prediction.
However, all these methods focused excessively on salient instances in training, which are easy instances with high confidence scores and can be easily classified. As a result, they overlook the importance of hard instances for training. In this paper, we intend to mine hard instances for improving WSI classification performance.

\subsection{Hard Sample Mining in Computer Vision}
Hard sample mining is a popular technique to speed up convergence and enhance the discriminative power of the model in many deep learning areas, such as face recognition~\cite{schroff2015facenet}, object detection~\cite{shrivastava2016training,wang2018hsm}, person re-identification~\cite{sun2019him_3,tan2022him_4,sheng2020hsm,ahmed2015improved}, and deep metric learning~\cite{suh2019him_2,sohn2016improved}. The main idea behind this technique is to select the samples which are hard to classify correctly (i.e., hard negatives and hard positives) for alleviating the imbalance between positive and negative samples and facilitating model training. 
There are generally three groups of approaches for evaluating sample difficulty: loss-based~\cite{hermans2017defense}, similarity-based~\cite{chen2017beyond}, and learnable weight-based~\cite{xu2019learning}. 
Typically, these strategies require complete sample supervision information. 
Drawing on the ideas of the above works, we propose a hard instance mining approach in MIL, mining hard examples at the instance level. 
In this, there are no complete instance labels, only the bag label is available.
Similar to our approach, Li~\etal utilized attention scores to identify salient instances from false negative bags to serve as hard negative instances and used them to compose the hard bags for improving classification performance~\cite{li2019hnm}. 
A key difference is that we indirectly mine hard instances by masking out the most salient instances rather than directly locating hard negative instances.

\begin{figure*}[t]
\centerline{\includegraphics[width=13.5cm]{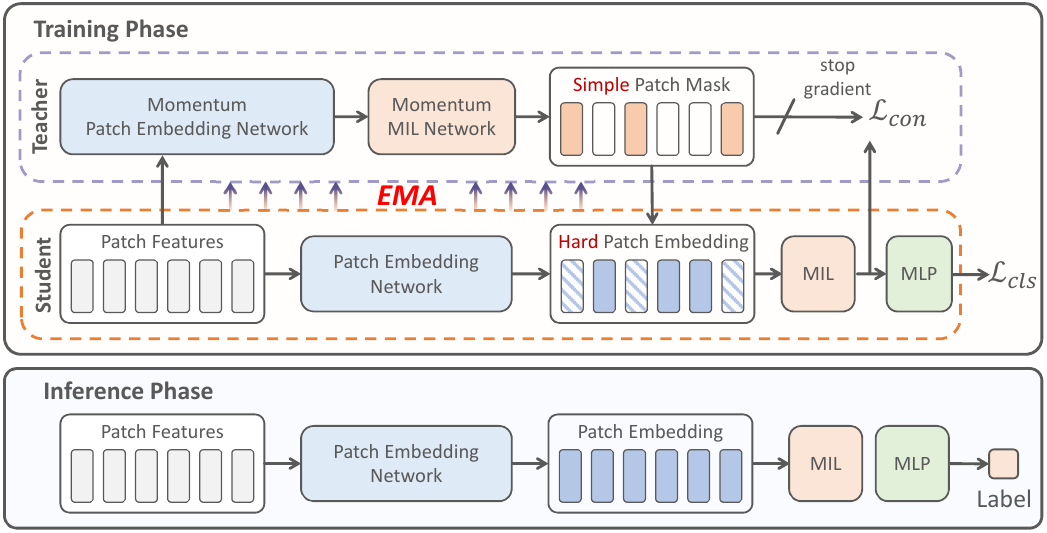}}
\caption{Overview of proposed MHIM-MIL. A momentum teacher is used to compute attention scores for all instances. We mask instances based on attention with hard mining strategies and feed the remaining to the student model. The student is updated by a consistency loss term $\mathcal{L}_{con}$ and a label error loss term $\mathcal{L}_{cls}$. The teacher parameters are updated with an Exponential Moving Average (EMA) of the student parameters without gradient updates. In the inference phase, we use the complete input instances and the student model only.
}
\label{fig:model}
\vspace{-0.3cm}
\end{figure*}

\section{Proposed Method}

\subsection{Background: MIL Formulation}
In MIL, any input WSI $X$ is considered as a bag with multiple instances, which can be represented as $X = \left\{x_i\right\}^N_{i=1}$. $x_i$ is a patch collected from the WSI and considered as the $i$-th instance of $X$. $N$ is the number of instances. For a classification task, there exists a known label $Y \in C$ for the bag and an unknown label $y_n \in C$ for each instance, where $C$ is the collection of category labels. The goal of a MIL model $\mathcal{M}(\cdot)$ is to predict the bag label with all instances $\hat{Y}\leftarrow\mathcal{M}(X)$. The popular solution is to learn a bag representation $F$ from the extracted features of instances $Z=\{z_i\}_{i=1}^N$ in a bag, which is also referred as the instance aggregation step. And a classifier $\mathcal{C}(\cdot)$, trained upon the $F$, can be used to predict the bag label $\hat{Y}\leftarrow\mathcal{C}(F)$. There are two ways to aggregate instances for achieving bag embedding. One is the attention-based aggregation~\cite{ilse2018attention} denoted as follows,
\begin{equation}
F = \sum_{i=1}^{N}a_{i}z_{i} \in \mathbb{R}^{D},
\end{equation}
where $a_{i}$ is the learnable scalar weight for $z_i$, and $D$ is the dimension of vector $F$ and $z_i$. Many works~\cite{li2021dual,clam,zhang2022dtfd} follow this formulation but differ in the ways they generate the attention score $a_i$.

Another is the multi-head self-attention (MSA) based aggregation~\cite{shao2021transmil}. In this fashion, a class token $z_{0}$ is embedded with the instance features to get the initial input sequence $ Z^0=\left[z_{0},z_1,\dots,z_N\right]\in \mathbb{R}^{\left(N+1\right) \times D}$ for aggregating instance features. This can be formulated as,
\begin{equation}
    \begin{aligned}
&\text{head} = A^{\ell}\left (Z^{\ell-1}W^{V}\right ) \in \mathbb{R}^{N\times \frac{D}{H}}, &\ell=1\dots L\\
&Z^{\ell} =  \textrm{Concat}\left ( \textrm{head}_{1},\cdots, \textrm{head}_{H}\right )W^O, &\ell=1\dots L
\end{aligned}
\end{equation}
where $W^{V}\in \mathbb{R}^{D \times \frac{D}{H}}$ and $W^{O}\in \mathbb{R}^{D \times D}$ are the learnable projection matrices of MSA. $A^{\ell}\in \mathbb{R}^{(N+1)\times (N+1)}$ is the attention matrix of the $\ell$-th layer, $L$ is the number of MSA block, and $H$ is the number of head in each MSA block. The bag embedding $F$ is the output class token at the final layer,
\begin{equation}
F = Z_{0}^{L}.
\end{equation}
The self-attention-based bag embedding is essentially a special case of attention-based bag embedding in the multi-instance learning setting. Collectively, these approaches can be referred to as the general attention-based MIL method.

\subsection{MHIM-MIL for WSI Classification}
In general attention-based MIL frameworks, the attention scores of instances indicate the contributions of instances to the bag classification. 
The salient instances with high scores are useful for classifying WSI in the testing phase but are not conducive to training a MIL model with good generalization ability. 
Although hard samples have been proven to enhance the generalization ability of the model in many computer vision scenarios~\cite{dong2017him_1,tan2022him_4,suh2019him_2,sun2019him_3},
previous MIL works focus more on exploiting the salient instances and neglecting the utilization of hard instances in model optimization.

In this paper,  we propose a simple and efficient MIL framework with Masked Hard Instance Mining (MHIM-MIL) to boost the WSI classification. 
As illustrated in Figure~\ref{fig:model}, the MHIM-MIL framework employs a Siamese structure during the training phase. The main component of our framework is a general attention-based MIL model (Student), denoted as $\mathcal{S(\cdot)}$, for aggregating instance features. To increase the discriminatory difficulty of the student model and force it to focus on hard instances, we introduce a momentum teacher, denoted as $\mathcal{T(\cdot)}$, to score the instances with attention weights and then employ some masked hard instance mining strategies to mask the salient instances while preserving the hard instances.
After hard instance mining, all the mined features are forwarded into the student model for the inference of the bag label.
The teacher shares the same network structure as the student model but does not need gradient-based updates. It is worth mentioning that, due to the varying number of instances within each bag, the non-batch gradient descent algorithm (i.e., SGD with batch size 1) is typically employed to optimize the MIL model. Therefore, compared to the traditional MIL frameworks with two-tier gradient updating models~\cite{zhang2022dtfd,xu2019camel}, this Siamese structure makes training more stable and efficient with fewer parameters. 
The proposed framework can be defined as,
\begin{equation}
    \hat{Y} = \mathcal{S}\left ( \hat{Z} \right ) = \mathcal{S}\left ( M_{\mathcal{T}} \left ( Z \right ) \right ),
\end{equation}
where $M_{\mathcal{T}}(\cdot)$ denotes a masked hard instance mining strategy through the teacher model and $\hat{Z}$ are the mined instances.

\subsection{Masked Hard Instance Mining Strategy}
Conventional hard sample mining strategies are difficult to apply without instance-level supervision. 
We address this challenge by proposing masked hard instance mining strategies that use attention scores to implicitly mine hard instances by masking out easy instances with high attention scores.
More specifically, given a complete sequence of instance features $Z =  \{ z_i\}^{N}_{i=1}$ as the input of the teacher model $\mathcal{T}\left ( \cdot  \right )$, the teacher outputs the attention weight $a_i$ for each instance as follow,
\begin{equation}
    A = \left [ a_1,\dots,a_i,\dots,a_N \right ] = \mathcal{T}\left ( Z  \right ).
\end{equation}
Then, we obtain the indices of the attention sequence in descending order by applying a sorting operation on $A$,
\begin{equation}
    I = \left [ i_{1},i_{2},\dots,i_{N} \right ] = \textrm{Sort}\left ( A  \right ),
\end{equation}
where $i_{1}$ is the index of the instance with the highest attention score while $i_N$ is the index of the one with the lowest score. With this index collection $I$, we will present several masked hard instance mining strategies to select the hard instances.
We define an $N$-dimensional binary vector $M=[m_1,\dots,m_i,\dots,m_N]$ for encoding the mask flags of instances where $m_i\in\{0,1\}$. If $m_i=1$, the $i$-th instance is masked, otherwise, it is unmasked.

\noindent\textbf{High Attention Masking:}
The simplest masked hard instance mining strategy is the High Attention Masking (HAM) strategy, which simply masks instances with the top $\beta_h$\% highest attention scores. The instance mask flags under HAM are initialized as all zero vectors, $M_{h}(:)=0$. Then we collect the indices of the instances whose scores are ranked in the top $\beta_h$\%, $I_h=[i_{t}]_{t=1}^{ \left \lceil \beta_h\% \times N \right \rceil}$. Finally, we set the mask flags with these indices, $M_h(I_h)=1$. 
To ensure that positive instances are preserved within the unmasked sequences, we also utilized techniques such as mask ratio decay. 

\begin{figure}[t]

    \begin{minipage}{1\linewidth}
    \centering
    \includegraphics[width=1\textwidth]{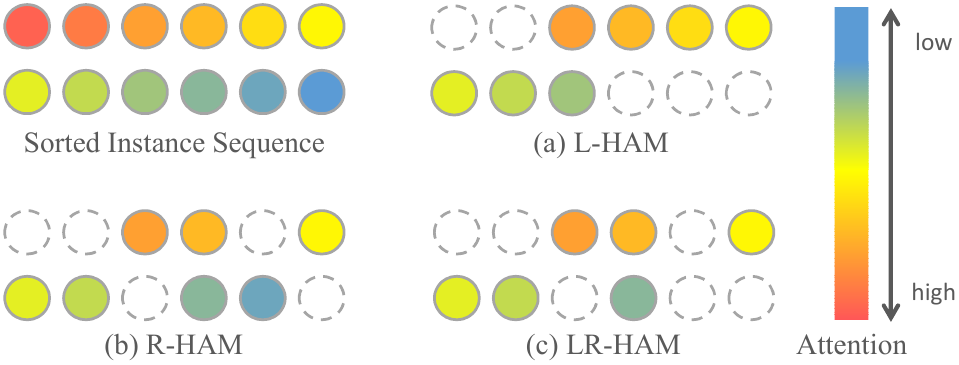}
    \end{minipage}
    \caption{Illustration of proposed hybrid masking strategy for hard instance mining.}
    \label{fig:mask_method}
\end{figure}

\noindent\textbf{Hybrid Masking:} We combine HAM with several other instance masking strategies as hybrid masking strategies to achieve some specific properties in hard instance mining, as shown in Figure~\ref{fig:mask_method}.
We consider the obtained mask flags as a collection and employ the union operation for mask flag fusion. We design three hybrid masking strategies as follows:
\begin{itemize}
    \item \textbf{L-HAM}: 
     We use the same pipeline as HAM to generate the mask flags $M_l$ for masking the instances with the top $\beta_l$\% lowest attention scores in order to filter out the redundant 
     uninformative instances and improve efficiency. 
     To endower this property to HAM, we union the mask flags obtained by two strategies to get the new mask flags, $\hat{M}=M_h \cup M_l$.
    \item \textbf{R-HAM}: Randomness is beneficial to reduce the risk of over-fitting. We generate a random mask flag vector $M_r$ with a given random ratio $\beta_r$\%, and combine it with $M_h$ for introducing the randomness to the hard instance mining, $\hat{M} = M_h \cup M_r $. 
    \item \textbf{LR-HAM}: Combining the above strategies, we can obtain completely hybrid mask flags,
    $\hat{M} = M_h \cup M_r \cup  M_l $, which is expected to achieve both of the mentioned desirable properties.
\end{itemize}
Once the final mask flag $\hat{M}$ is produced, the masked instance sequence will be obtained:
\begin{equation}
    \hat{Z} = M_{\mathcal{T}} \left ( Z \right ) = \textrm{Mask}\left ( Z,\hat{M} \right ) \in \mathbb{R}^{\hat{N}\times D},
\end{equation}
where the $\hat{N}$ is the number of unmasked instances.

\subsection{Consistency-based Iterative Optimization}
Under the Siamese structure, while the teacher model guides the training of the student model, the new knowledge learned by the student model will also update the teacher model.
This iterative optimization process progressively improves the mining ability of the teacher and the discriminability of the student. To further facilitate this optimization and explore additional supervised information provided by the momentum teacher, we propose a consistency loss that constrains the classification results of both models. 

\noindent\textbf{Student Optimization:}
 There are two losses in student optimization. One is the cross-entropy for measuring the bag label prediction loss,
\begin{equation}
   \mathcal{L}_{cls} = Y \textrm{log}\hat{Y} + \left ( 1-Y \right )\textrm{log}\left ( 1-\hat{Y} \right ).
\end{equation}
Another is a consistency loss between the bag representation of student $F_s$ and momentum teacher $F_t$,
\begin{equation}
   \mathcal{L}_{con} = - \textrm{softmax}\left ( F_t / \tau  \right ) \, \textrm{log}\, F_s
\end{equation}
where the $\tau > 0$ is a temperature parameter.
Overall, the final optimization loss is as follows:
\begin{equation}
    \begin{aligned}
   \{\hat{\theta_s}\}\leftarrow \arg\underset{\theta_s}\min ~\mathcal{L} = \mathcal{L}_{cls}+ \alpha \mathcal{L}_{con}
    \end{aligned}
\end{equation}
where  $\theta_s$ is the parameters of $\mathcal{S}(\cdot)$, and $\alpha$ is scaling factor.

\noindent\textbf{Teacher Optimization:}
The parameters of momentum teacher $\theta_t$ are updated by an exponential moving average (EMA) of the student parameters.
The update rule is $\theta_t \leftarrow \lambda \theta_t+(1-\lambda) \theta_s $, where $\lambda$ is a hyperparameter. More importantly, the updated teacher is utilized in the next iteration of hard instance mining.

\section{Experiments and Results}
\subsection{Datasets and Evaluation Metrics}
\textbf{CAMELYON-16~\cite{bejnordi2017diagnostic}} is a WSI dataset proposed for metastasis detection in breast cancer.
The dataset contains a total of 400 WSIs, which are officially split into 270 for training and 130
for testing, and the testing sample ratio is 13/40$ \approx$1/3.
Following~\cite{Zhang_2022_BMVC,clam,chen2022hipt},
we adopt 3-times 3-fold cross-validation on this dataset to ensure that each slide is used in training and testing, which can alleviate the impact of data split and random seed on the model evaluation. 
Each fold has approximately 133 slides. We report the mean and standard
deviation of performance metrics over 3 runs.

\textbf{TCGA Lung Cancer} includes two sub-type of cancers, Lung Adenocarcinoma (LUAD) and Lung Squamous Cell Carcinoma (LUSC). There are diagnostic slides, LUAD with 541 slides from 478 cases, and LUSC with 512 slides from 478 cases. We randomly split the dataset into training, validation, and testing sets with a ratio of 65:10:25 on the patient level. 4-fold cross-validation is adopted, and the mean and standard
deviation of performance metrics of the 4 test folders are reported.

We adopt the same data pre-processing as in the CLAM~\cite{clam}.
Following the previous work~\cite{shao2021transmil, clam} we leverage Accuracy, Area Under Curve (AUC), and F1-score to evaluate model performance.
AUC is the primary performance metric in the binary classification task, and we only report AUC in ablation experiments. Please refer to the \textbf{Supplementary Material} for the details of these two datasets.


\begin{table*}[!h]
\small
\centering
\begin{tabular}{lcccccc}
\toprule
{\multirow{2}{*}{Method}} & \multicolumn{3}{c}{CAMELYON-16} & \multicolumn{3}{c}{TCGA Lung Cancer} \\ \cmidrule(lr){2-4} \cmidrule(lr){5-7}
                          & Accuracy       & AUC           & F1-score       & Accuracy       & AUC            & F1-score      \\\midrule
Max-pooling                      & 78.95$\pm2.28$              & 81.28$\pm3.74$               & 71.06$\pm2.59$              
                                 & 81.49$\pm$1.24              & 86.45$\pm$0.71             & 80.56$\pm$1.09            \\
Mean-pooling                     & 76.69$\pm0.20$              & 80.07$\pm0.78$             & 70.41$\pm0.16$              
                                 & 84.14$\pm$2.97              & 90.13$\pm$2.40             & 83.39$\pm$3.14             \\
AB-MIL~\cite{ilse2018attention}  & 90.06$\pm$0.60              & 94.00$\pm$0.83             & 87.40$\pm$1.05               
                                 & 88.03$\pm$2.19              & 93.17$\pm$2.05             & 87.41$\pm$2.42         \\
DSMIL~\cite{li2021dual}          & 90.17$\pm$1.02             & 94.57$\pm$0.40           & 87.65$\pm$1.18

                                 & 88.32$\pm$2.70              & 93.71$\pm$1.82             & 87.90$\pm$2.50 \\
CLAM-SB~\cite{clam}              & 90.31$\pm$0.12              & 94.65$\pm$0.30             & 87.89$\pm$0.59              
                                 & 87.74$\pm$2.22              & 93.67$\pm$1.64             & 87.36$\pm$2.24\\
CLAM-MB~\cite{clam}              &  90.14$\pm$0.85              & 94.70$\pm$0.76               & 88.10$\pm$0.63            
                                 & 88.73$\pm$1.62              & 93.69$\pm$0.54             & 88.28$\pm$1.58   \\
TransMIL~\cite{shao2021transmil} & 89.22$\pm$2.32              & 93.51$\pm$2.13             & 85.10$\pm$4.33              
                                 & 87.08$\pm$1.97              & 92.51$\pm$1.76             & 86.40$\pm$2.08  \\
DTFD-MIL~\cite{zhang2022dtfd}    & 90.22$\pm$0.36              &  95.15$\pm$0.14            & 87.62$\pm$0.59             
                                 & 88.23$\pm$2.12              & 93.83$\pm$1.39             & 87.71$\pm$2.04  \\

\rowcolor{dino}MHIM-MIL (AB-MIL)        & 91.81$\pm$0.82     & 96.14$\pm$0.52      & 89.94$\pm$0.70      
                                 & 89.64$\pm$2.25 & 94.97$\pm$1.72 & 89.31$\pm$2.19 \\
\rowcolor{dino}MHIM-MIL (TransMIL)         & 91.98$\pm$0.89     & \textbf{96.49$\pm$0.48}      & 90.13$\pm$1.08      
                                 & \textbf{90.02$\pm$2.59} & 94.87$\pm$2.17 & 89.65$\pm$2.63 \\
\rowcolor{dino}MHIM-MIL (DSMIL)         & \textbf{92.48$\pm$0.35}     & 96.49$\pm$0.65      & \textbf{90.75$\pm$0.73}      
                                 & 89.83$\pm$3.37 & \textbf{95.53$\pm$1.74} & \textbf{89.71$\pm$2.92} \\
\bottomrule
\end{tabular}
\caption{The performance of different MIL approaches on CAMELYON-16 (C16) and TCGA Lung Cancer (TCGA). The highest performance is in bold. The Accuracy and F1-score are determined by the optimal threshold. }
\vspace{-0.2cm}
\label{tab:main}
\end{table*}
\subsection{Implementation Details}
The details on network architectures and training are described in \textbf{Supplementary Material}.
\subsection{Performance Comparison with Exiting Works}
We mainly compare with AB-MIL~\cite{ilse2018attention}, DSMIL~\cite{li2021dual}, CLAM-SB~\cite{clam}, CLAM-MB~\cite{clam}, TransMIL~\cite{shao2021transmil}, and DTFD-MIL~\cite{zhang2022dtfd}, all of which are attention-based MIL methods. 
In addition, we compared two traditional MIL pooling operations, Max-pooling and Mean-pooling. 
Due to the dataset differences, the results of all other methods are reproduced using the official code they provide under the same settings.

As shown in Table~\ref{tab:main}, max-pooling and mean-pooling perform poorly on two datasets compared to other methods. We attribute this to their insufficient modeling of key instance information. Simple pooling operations are prone to be misled by limited slides that contain numerous instances. This problem is especially severe on the CAMELYON-16 dataset, where the proportion of significant instances is extremely small. For example, max-pooling lags behind DTFD-MIL~\cite{zhang2022dtfd} by 13.87\% on AUC. Attention-based methods achieve better performance on both datasets by focusing on salient instances. 
In particular, the representative MIL framework DTFD-MIL~\cite{zhang2022dtfd} benefits from the further exploration of significant instances and achieves the second-best performance on both datasets (95.15\% AUC on CAMELYON-16 and 93.83\% AUC on TCGA). 
However, it also suffers from overemphasizing salient instances during training, which limits its generalization.
Our proposed MHIM-MIL achieves significant performance improvement on both datasets (+1.34\% AUC on CAMELYON-16 and +1.70\% AUC on TCGA) by mining hard instances during training, breaking the performance bottleneck.
It is worth mentioning that we validate our framework on three representative MIL models, both of which can outperform the existing MIL methods. 

\subsection{Computational Cost Analysis}
\begin{table}[tb]
\small
\centering
\begin{tabular}{lccccc}
\toprule
   Model           & C16 & TCGA &Para. & Time & Mem.\\ \midrule
AB-MIL & 94.00 & 93.17 &657K & \textbf{4.0s} & 2.4G \\ 
CLAM-MB & 94.70 & 93.69 &789K & 4.3s & 2.7G \\
DTFD-MIL & 95.15 & 93.83 &987K & 5.2s & \textbf{2.1G} \\
\rowcolor{dino}MHIM-MIL & \textbf{96.14} & \textbf{94.97} &\textbf{657K} &  4.3s & 2.3G \\
\midrule
TransMIL & 93.51 & 92.51 &2.67M & 13.1s & 10.6G \\
\rowcolor{dino}MHIM-MIL & \textbf{96.49} & \textbf{94.87} &\textbf{2.67M} & \textbf{10.1s} & \textbf{5.5G} \\
\bottomrule
\end{tabular}
\caption{Comparison of time and memory requirements of different MIL methods. We report the model size (Para.), the training time per epoch (Time), and the peak memory usage (Mem.) on the CAMELYON-16 dataset (C16).}
\vspace{-0.3cm}
\label{tab:effi}
\end{table}

In this section, we report the training time and GPU memory requirements for running different MIL models on a 3090 GPU. The upper part of Table~\ref{tab:effi} compares some MIL frameworks that use AB-MIL~\cite{ilse2018attention} as a baseline. We observe that traditional MIL frameworks typically introduce additional parameters and reduce efficiency due to their complex structures. For example, the state-of-the-art framework DTFD-MIL~\cite{zhang2022dtfd} increases the parameter size by nearly twice (657K vs. 987K) and the training time by 30\%. In contrast, MHIM-MIL achieves the most significant performance improvement with almost no extra computational cost due to the momentum teacher. Moreover, existing Transformer-based MIL methods are usually plagued by high computing costs due to their large number of parameters and self-attention operations.
For instance, TransMIL~\cite{shao2021transmil}, which first applies a pure Transformer MIL model to solve WSI classification problems, has 4$\times$ more parameters than AB-MIL, 3$\times$ longer training time, and almost 4.5$\times$ higher memory consumption.
Furthermore, the extremely long input sequences in WSI classification degrade the stability of such complex structures (2.13\% AUC standard deviation on C16, which is the highest among all embedding-level MIL methods). With the masked hard instance mining strategy, the MHIM-MIL framework significantly reduces the computational cost (-24\% training time and -48\% memory usage) and enhances its stability (0.48\% AUC standard deviation on C16). More details are provided in \textbf{Supplementary Material}.

\begin{table}[tbp]
\small
\centering
    \begin{tabularx}{0.99\linewidth}{lcccc}
\toprule
{\multirow{2}{*}{Module}} & \multicolumn{2}{c}{CAMELYON-16} & \multicolumn{2}{c}{TCGA} \\ \cmidrule(lr){2-3} \cmidrule(lr){4-5}
                          & AB.         & Trans.        & AB.     & Trans.     \\\midrule
Baseline                  & 94.00         & 93.51           & 93.17     & 92.51        \\
+MHIM                    & 95.86         & 96.06           & 94.14     & 93.75 \\
+MHIM+Siam.                     & 95.82         & 96.24           & 94.55     & 94.13        \\
\rowcolor{dino}+MHIM+Siam.+Con.                  & \textbf{96.14}         & \textbf{96.49}           & \textbf{94.97}     & \textbf{94.87} \\   
\bottomrule 
\end{tabularx}

\caption{The effect of different components in MHIM-MIL with two MIL models: AB-MIL (AB.) and TransMIL (Trans.). MHIM denotes the masked hard instance mining strategy. Siam. refers to the Siamese framework. Con. represents consistency loss.}
\label{tab:abl}
\end{table}

\subsection{Ablation Study}

\subsubsection{Importance of the Different Components}
Table~\ref{tab:abl} shows the effect of different modules in MHIM-MIL on two datasets. The baseline methods are two representative attention-based MIL methods, namely AB-MIL~\cite{ilse2018attention} and TransMIL~\cite{shao2021transmil}. 
First, we introduce the naive masked hard instance mining strategy, which leverages the model itself to mine hard instances during training. 
This strategy improves AUC by 1.86\% and 2.55\% for the two MIL models on CAMELYON-16 respectively, indicating that concentrating on hard instances during training can assist mainstream MIL models in building better classification boundaries.
Further discussion on the masked hard instance mining strategy is presented in Section~\ref{sec:diff_hil}. Compared with the naive MHIM strategy, the third row of the table suggests that a Siamese structure~\cite{caron2021dino,bromley1993siamese,chen2021simsiam} based on a momentum teacher is beneficial for more stable and effective mining of hard instances. 
We elaborate more on choosing the teacher model in Section~\ref{sec:diff_tea}. 
After adding consistency loss term to the objective function, our full MHIM-MIL framework achieves the best performance (96.49\% AUC on CAMELYON-16 and 94 .97\% AUC on TCGA). 
For subsequent ablation experiments, we include consistency loss by default to facilitate the optimization of our framework.

\begin{table}[tb]
\small
\centering
    \begin{tabularx}{0.87\linewidth}{p{1.8cm}cccc}
\toprule
{\multirow{2}{*}{Strategy}}  & \multicolumn{2}{c}{CAMELYON-16} & \multicolumn{2}{c}{TCGA} \\ \cmidrule(lr){2-3} \cmidrule(lr){4-5}
                & AB.         & Trans.        & AB.     & Trans.     \\\midrule
Baseline                  & 94.00         & 93.51           & 93.17     & 92.51        \\
HAM               & 95.68        &95.90      & 93.83  &94.54 \\
\rowcolor{dino}R-HAM             & \textbf{96.14}        &95.88     & 94.79  &94.60  \\ 
\rowcolor{dino}L-HAM             & 95.81        &\textbf{96.49}     & 94.33  &94.67\\
\rowcolor{dino}LR-HAM           & 95.92        &96.33     & \textbf{94.97} &\textbf{94.87}\\ 

\bottomrule 
\end{tabularx}
\caption{Comparison between different masked hard instance mining strategies. The three hybrid strategies show varying performance across the benchmarks.}
\vspace{-0.3cm}
\label{tab:mask}
\end{table}

\subsubsection{Impact of the Different MHIM Strategies}
\label{sec:diff_hil}
The masked hard instance mining strategy is the core design of our method. The main idea of this strategy is masking the most salient instances to indirectly mine hard instances to facilitate model training. Based on this idea, we devise three hybrid strategies (R-HAM, L-HAM, and LR-HAM) and present their impact in Table~\ref{tab:mask}. The basic strategy, High Attention Masking (HAM), already boosts performance significantly, leading to AUC improvements of 1.68\% and 2.39\% for two MIL models on the CAMELYON-16 dataset, respectively. 
After introducing the other two strategies, different MIL models achieve performance improvements on both datasets. 
Specifically, AB-MIL~\cite{ilse2018attention} shows more significant performance gains after introducing randomness (96.14\% AUC on CAMELYON-16 with R-HAM) due to its better ability to filter out redundant information, while TransMIL~\cite{shao2021transmil} shows the reverse trend (96.49\% AUC on CAMELYON-16 with L-HAM). Furthermore, the more complex three-hybrid strategy (LR-HAM) achieves the best performance on the TCGA dataset, which has a larger proportion of positive areas and more instances. 
Overall, our experiments validate the effectiveness of masked hard instance mining strategy, and the diversity of proposed strategies improves its applicability to different datasets and MIL models.

\begin{table}[tbp]
\small
\centering
    \begin{tabularx}{0.9\linewidth}{lcccc}
\toprule
{\multirow{2}{*}{Teacher}} & \multicolumn{2}{c}{CAMELYON-16} & \multicolumn{2}{c}{TCGA} \\ \cmidrule(lr){2-3} \cmidrule(lr){4-5}
                          & AB.         & Trans.        & AB.     & Trans.     \\\midrule
Baseline                  & 94.00         & 93.51           & 93.17     & 92.51        \\
Student copy                      & 95.84         & 95.86           & 93.68     & 93.45 \\
Init.                      & 95.88         & 96.12           & 94.66     & 94.15        \\   
Momentum                      & 95.96         & 96.11           & 94.65     & 94.45  \\
\rowcolor{dino}Init.+Momentum                & \textbf{96.14}         & \textbf{96.49}           & \textbf{94.97}     & \textbf{94.87} \\
\bottomrule 
\end{tabularx}
\includegraphics[width=0.7\linewidth]{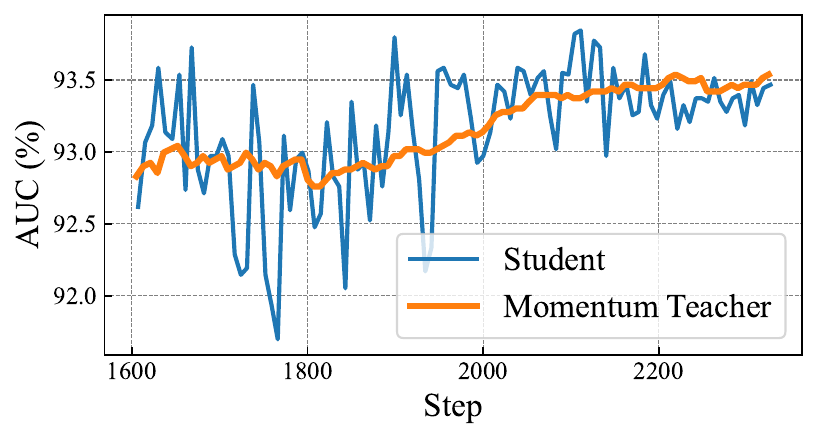}
\vspace{-0.5cm}
\caption{Comparison of different types of teachers. Momentum denotes the teacher is updated by EMA strategy. Init. indicates the initialization of the teacher with pre-trained parameters. The bottom figure compares the stability of the momentum teacher and the non-batch gradient updated student during training.}
\label{tab:tea}
\end{table}

\begin{figure*}[t]
\centering
    \includegraphics[width=\linewidth]{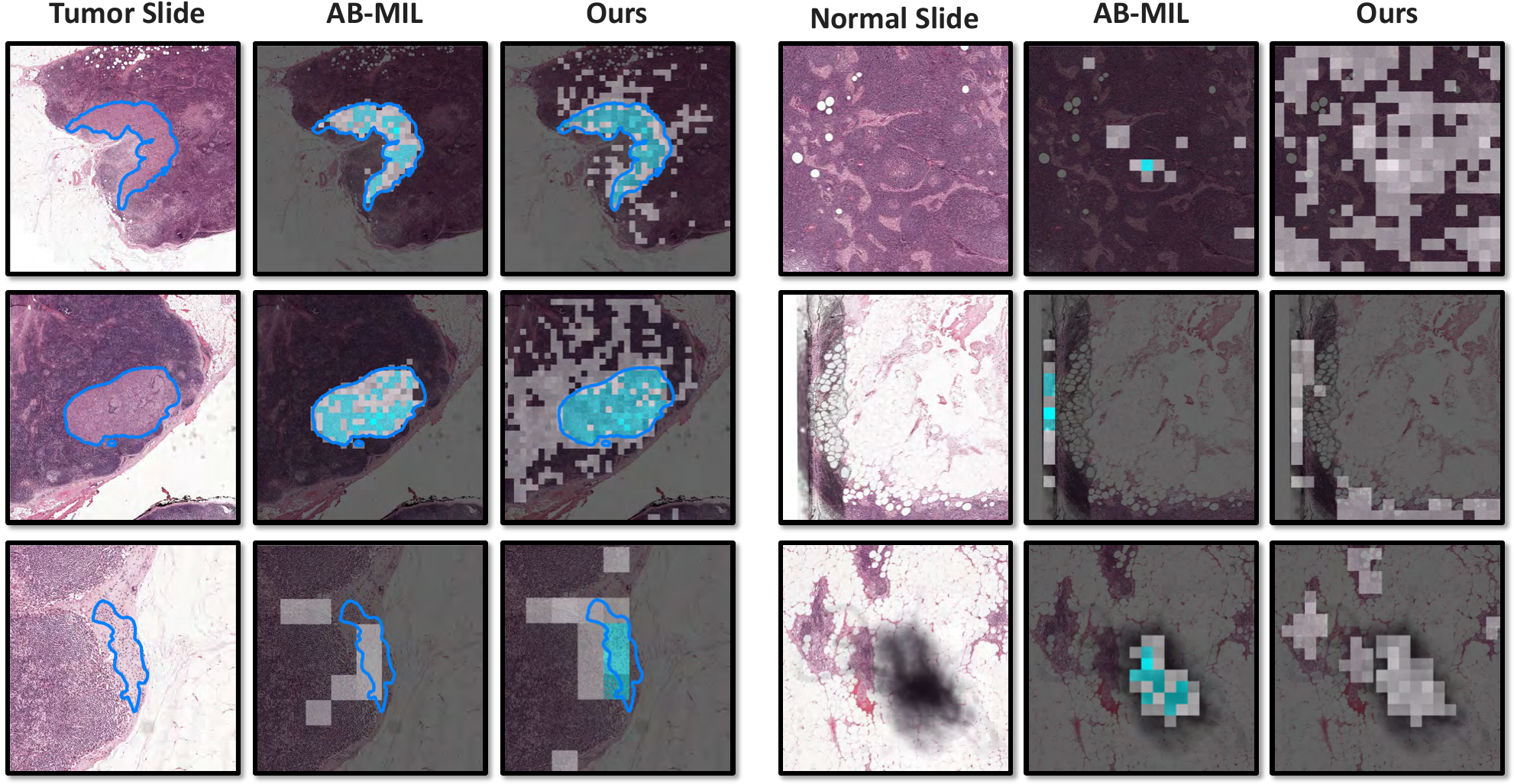}
    \caption{Patch visualization produced by AB-MIL~\cite{ilse2018attention} (baseline) and MHIM-MIL. The \textcolor{blue}{blue} lines outline the tumor regions. The brighter patch indicates higher attention scores. The \textcolor{cyan}{cyan} colors indicate high probabilities of being tumor for the corresponding locations. Ideally, the \textcolor{cyan}{cyan} patches should cover only the area within the \textcolor{blue}{blue} lines. We show that focusing only on more salient regions reduces the generalization ability of the model and that hard instances can provide useful information for more accurate and comprehensive judgments.} 
    \vspace{-0.2cm}
    \label{fig:vis}
\end{figure*}

\subsubsection{Impact of the Choice of Teacher Network}
\label{sec:diff_tea}
In MHIM-MIL, we employ a Teacher model to mine hard instances and facilitate training of the Student model. In Table~\ref{tab:tea}, we comprehensively investigate the effects of various choices of Teacher network.  
First, we utilize a single-model structure, which treats the Student model as the Teacher.
The student conducts masked hard instance mining prior to training. 
Due to the non-batch gradient update, the unstable performance of the Student model makes the strategy susceptible to noise, so the performance is not optimal.
Second, we adopt a momentum teacher, which shares the same network structure as the Student model and is updated with the EMA strategy.
This updating strategy enhanced the stability of momentum teachers, as shown in the figure below, and enabled MHIM-MIL to achieve 0.97\% and 1.00\% performance improvement in TCGA under the two MIL models, respectively.
With proper initialization, the momentum teacher achieves the best performance. However, a fixed initialization teacher fails to learn new knowledge, which emphasizes the significance of iterative optimization.

\subsection{Visualization} 
To more intuitively understand the effect of the masked hard instance mining, we visualize the attention scores (bright patch) and tumor probabilities (\textcolor{cyan}{cyan} patch) of patches produced by AB-MIL and MHIM-MIL, as illustrated in Figure~\ref{fig:vis}.
Here, MHIM-MIL employs AB-MIL as its baseline model. 
We note that attention scores only indicate the regions of interest of models and are infeasible to reflect tumor probabilities~\cite{zhang2022dtfd,li2021dual}. 
First, as shown in Figure~\ref{fig:vis}, AB-MIL often assigns high tumor probabilities to patches in non-tumor areas. We attribute this phenomenon to the low generalization capability of conventional attention-based MIL models, which tend to focus only on salient regions during training. In contrast, MHIM-MIL trained with hard instances shows a much better generalization ability than the baseline model for noise robustness (rows \textcolor{dino_text}{2} and \textcolor{dino_text}{3} on the right) and for precise detection of challenging subtle tumor areas (row \textcolor{dino_text}{3} on the left). 
More significantly, we find that focusing only on tumor areas leads to missing most of them, expanding the view to include some ``irrelevant areas” enables the model to make more complete judgments (rows \textcolor{dino_text}{1} and \textcolor{dino_text}{2} on the left). 
This phenomenon demonstrates how hard instances provide more useful information to help the model make more accurate and comprehensive judgments. 
We provide more details and an in-depth analysis of this patch visualization in \textbf{Supplementary Material}.

\section{Conclusion}

This paper rethinks the impact of salient instances for MIL-based WSI classification algorithms. We demonstrate that attention-based MIL methods excessively prioritizing salient instances harm the generalization ability of the model. 
To address this issue, we have proposed several masked hard instance mining strategies that mask out salient patches and encourage the model to attend to informative regions for better discriminative learning.
Through qualitative analysis, we have demonstrated that these strategies effectively alleviate the under-fitting problem of general AB-MIL to hard instances. 
We have also developed the MHIM-MIL framework that leverages momentum teacher and consistency loss to further enhance hard instance mining. 
Our experimental results demonstrate the superiority and generality of the MHIM-MIL framework over other latest methods. 
In future work, we plan to devise a more precise localization scheme for hard instances that can facilitate model training and convergence.

\section{Acknowledgement}
Reported research is partly supported by the National Natural Science Foundation of China under Grant 62176030, and the Natural Science Foundation of Chongqing under Grant cstc2021jcyj-msxmX0568.

{\small
\bibliographystyle{ieee_fullname}
\bibliography{egbib}

\begin{thebibliography}{10}\itemsep=-1pt

\bibitem{ahmed2015improved}
Ejaz Ahmed, Michael Jones, and Tim~K Marks.
\newblock An improved deep learning architecture for person re-identification.
\newblock In {\em Proceedings of the IEEE conference on computer vision and
  pattern recognition}, pages 3908--3916, 2015.

\bibitem{bejnordi2017diagnostic}
Babak~Ehteshami Bejnordi, Mitko Veta, Paul~Johannes Van~Diest, Bram
  Van~Ginneken, Nico Karssemeijer, Geert Litjens, Jeroen~AWM Van Der~Laak,
  Meyke Hermsen, Quirine~F Manson, Maschenka Balkenhol, et~al.
\newblock Diagnostic assessment of deep learning algorithms for detection of
  lymph node metastases in women with breast cancer.
\newblock {\em JAMA}, 318(22):2199--2210, 2017.

\bibitem{bromley1993siamese}
Jane Bromley, Isabelle Guyon, Yann LeCun, Eduard S{\"a}ckinger, and Roopak
  Shah.
\newblock Signature verification using a" siamese" time delay neural network.
\newblock {\em Advances in neural information processing systems}, 6, 1993.

\bibitem{campanella2019clinical}
Gabriele Campanella, Matthew~G Hanna, Luke Geneslaw, Allen Miraflor, Vitor
  Werneck~Krauss Silva, Klaus~J Busam, Edi Brogi, Victor~E Reuter, David~S
  Klimstra, and Thomas~J Fuchs.
\newblock Clinical-grade computational pathology using weakly supervised deep
  learning on whole slide images.
\newblock {\em Nature Medicine}, 25(8):1301--1309, 2019.

\bibitem{caron2021dino}
Mathilde Caron, Hugo Touvron, Ishan Misra, Herv{\'e} J{\'e}gou, Julien Mairal,
  Piotr Bojanowski, and Armand Joulin.
\newblock Emerging properties in self-supervised vision transformers.
\newblock In {\em Proceedings of the IEEE/CVF international conference on
  computer vision}, pages 9650--9660, 2021.

\bibitem{chen2022hipt}
Richard~J Chen, Chengkuan Chen, Yicong Li, Tiffany~Y Chen, Andrew~D Trister,
  Rahul~G Krishnan, and Faisal Mahmood.
\newblock Scaling vision transformers to gigapixel images via hierarchical
  self-supervised learning.
\newblock In {\em Proceedings of the IEEE/CVF Conference on Computer Vision and
  Pattern Recognition}, pages 16144--16155, 2022.

\bibitem{chen2017beyond}
Weihua Chen, Xiaotang Chen, Jianguo Zhang, and Kaiqi Huang.
\newblock Beyond triplet loss: a deep quadruplet network for person
  re-identification.
\newblock In {\em Proceedings of the IEEE conference on computer vision and
  pattern recognition}, pages 403--412, 2017.

\bibitem{chen2021simsiam}
Xinlei Chen and Kaiming He.
\newblock Exploring simple siamese representation learning.
\newblock In {\em Proceedings of the IEEE/CVF conference on computer vision and
  pattern recognition}, pages 15750--15758, 2021.

\bibitem{chikontwe2021dual}
Philip Chikontwe, Miguel Luna, Myeongkyun Kang, Kyung~Soo Hong, June~Hong Ahn,
  and Sang~Hyun Park.
\newblock Dual attention multiple instance learning with unsupervised
  complementary loss for covid-19 screening.
\newblock {\em Medical Image Analysis}, 72:102105, 2021.

\bibitem{deng2009imagenet}
Jia Deng, Wei Dong, Richard Socher, Li-Jia Li, Kai Li, and Li Fei-Fei.
\newblock Imagenet: A large-scale hierarchical image database.
\newblock In {\em 2009 IEEE conference on computer vision and pattern
  recognition}, pages 248--255. Ieee, 2009.

\bibitem{dietterich1997mil_2}
Thomas~G Dietterich, Richard~H Lathrop, and Tom{\'a}s Lozano-P{\'e}rez.
\newblock Solving the multiple instance problem with axis-parallel rectangles.
\newblock {\em Artificial intelligence}, 89(1-2):31--71, 1997.

\bibitem{dietterich1997solving}
Thomas~G Dietterich, Richard~H Lathrop, and Tom{\'a}s Lozano-P{\'e}rez.
\newblock Solving the multiple instance problem with axis-parallel rectangles.
\newblock {\em Artificial Intelligence}, 89(1-2):31--71, 1997.

\bibitem{dong2017him_1}
Qi Dong, Shaogang Gong, and Xiatian Zhu.
\newblock Class rectification hard mining for imbalanced deep learning.
\newblock In {\em Proceedings of the IEEE international conference on computer
  vision}, pages 1851--1860, 2017.

\bibitem{instance_mil_1}
Ji Feng and Zhi-Hua Zhou.
\newblock Deep miml network.
\newblock In {\em Proceedings of the AAAI conference on artificial
  intelligence}, volume~31, 2017.

\bibitem{he2022transfg}
Ju He, Jie-Neng Chen, Shuai Liu, Adam Kortylewski, Cheng Yang, Yutong Bai, and
  Changhu Wang.
\newblock Transfg: A transformer architecture for fine-grained recognition.
\newblock In {\em Proceedings of the AAAI Conference on Artificial
  Intelligence}, volume~36, pages 852--860, 2022.

\bibitem{he2016deep}
Kaiming He, Xiangyu Zhang, Shaoqing Ren, and Jian Sun.
\newblock Deep residual learning for image recognition.
\newblock In {\em Proceedings of the IEEE conference on computer vision and
  pattern recognition}, pages 770--778, 2016.

\bibitem{hearst1998svm}
Marti~A. Hearst, Susan~T Dumais, Edgar Osuna, John Platt, and Bernhard
  Scholkopf.
\newblock Support vector machines.
\newblock {\em IEEE Intelligent Systems and their applications}, 13(4):18--28,
  1998.

\bibitem{hermans2017defense}
Alexander Hermans, Lucas Beyer, and Bastian Leibe.
\newblock In defense of the triplet loss for person re-identification.
\newblock {\em arXiv preprint arXiv:1703.07737}, 2017.

\bibitem{hou2016patch}
Le Hou, Dimitris Samaras, Tahsin~M Kurc, Yi Gao, James~E Davis, and Joel~H
  Saltz.
\newblock Patch-based convolutional neural network for whole slide tissue image
  classification.
\newblock In {\em Proceedings of the IEEE conference on computer vision and
  pattern recognition}, pages 2424--2433, 2016.

\bibitem{ilse2018attention}
Maximilian Ilse, Jakub Tomczak, and Max Welling.
\newblock Attention-based deep multiple instance learning.
\newblock In {\em International conference on machine learning}, pages
  2127--2136. PMLR, 2018.

\bibitem{kingma2014adam}
Diederik~P Kingma and Jimmy Ba.
\newblock Adam: A method for stochastic optimization.
\newblock {\em arXiv preprint arXiv:1412.6980}, 2014.

\bibitem{li2021dual}
Bin Li, Yin Li, and Kevin~W Eliceiri.
\newblock Dual-stream multiple instance learning network for whole slide image
  classification with self-supervised contrastive learning.
\newblock In {\em CVPR}, pages 14318--14328, 2021.

\bibitem{li2021dt}
Hang Li, Fan Yang, Yu Zhao, Xiaohan Xing, Jun Zhang, Mingxuan Gao, Junzhou
  Huang, Liansheng Wang, and Jianhua Yao.
\newblock Dt-mil: Deformable transformer for multi-instance learning on
  histopathological image.
\newblock In {\em International Conference on Medical Image Computing and
  Computer-Assisted Intervention}, pages 206--216. Springer, 2021.

\bibitem{li2019hnm}
Meng Li, Lin Wu, Arnold Wiliem, Kun Zhao, Teng Zhang, and Brian Lovell.
\newblock Deep instance-level hard negative mining model for histopathology
  images.
\newblock In {\em Medical Image Computing and Computer Assisted
  Intervention--MICCAI 2019: 22nd International Conference, Shenzhen, China,
  October 13--17, 2019, Proceedings, Part I 22}, pages 514--522. Springer,
  2019.

\bibitem{lu2021nature}
Ming~Y Lu, Tiffany~Y Chen, Drew~FK Williamson, Melissa Zhao, Maha Shady, Jana
  Lipkova, and Faisal Mahmood.
\newblock Ai-based pathology predicts origins for cancers of unknown primary.
\newblock {\em Nature}, 594(7861):106--110, 2021.

\bibitem{clam}
Ming~Y Lu, Drew~FK Williamson, Tiffany~Y Chen, Richard~J Chen, Matteo Barbieri,
  and Faisal Mahmood.
\newblock Data-efficient and weakly supervised computational pathology on
  whole-slide images.
\newblock {\em Nature Biomedical Engineering}, 5(6):555--570, 2021.

\bibitem{manivannan2017subcategory}
Siyamalan Manivannan, Caroline Cobb, Stephen Burgess, and Emanuele Trucco.
\newblock Subcategory classifiers for multiple-instance learning and its
  application to retinal nerve fiber layer visibility classification.
\newblock {\em IEEE Transactions on Medical Imaging}, 36(5):1140--1150, 2017.

\bibitem{maron1997mil_1}
Oded Maron and Tom{\'a}s Lozano-P{\'e}rez.
\newblock A framework for multiple-instance learning.
\newblock {\em Advances in neural information processing systems}, 10, 1997.

\bibitem{pinckaers2020intro_pami}
Hans Pinckaers, Bram Van~Ginneken, and Geert Litjens.
\newblock Streaming convolutional neural networks for end-to-end learning with
  multi-megapixel images.
\newblock {\em IEEE transactions on pattern analysis and machine intelligence},
  44(3):1581--1590, 2020.

\bibitem{schroff2015facenet}
Florian Schroff, Dmitry Kalenichenko, and James Philbin.
\newblock Facenet: A unified embedding for face recognition and clustering.
\newblock In {\em Proceedings of the IEEE conference on computer vision and
  pattern recognition}, pages 815--823, 2015.

\bibitem{shao2021transmil}
Zhuchen Shao, Hao Bian, Yang Chen, Yifeng Wang, Jian Zhang, Xiangyang Ji,
  et~al.
\newblock Transmil: Transformer based correlated multiple instance learning for
  whole slide image classification.
\newblock {\em NeurIPS}, 34, 2021.

\bibitem{sharma2021cluster}
Yash Sharma, Aman Shrivastava, Lubaina Ehsan, Christopher~A Moskaluk, Sana
  Syed, and Donald~E Brown.
\newblock Cluster-to-conquer: A framework for end-to-end multi-instance
  learning for whole slide image classification.
\newblock {\em arXiv preprint arXiv:2103.10626}, 2021.

\bibitem{sheng2020hsm}
Hao Sheng, Yanwei Zheng, Wei Ke, Dongxiao Yu, Xiuzhen Cheng, Weifeng Lyu, and
  Zhang Xiong.
\newblock Mining hard samples globally and efficiently for person
  reidentification.
\newblock {\em IEEE Internet of Things Journal}, 7(10):9611--9622, 2020.

\bibitem{shrivastava2016training}
Abhinav Shrivastava, Abhinav Gupta, and Ross Girshick.
\newblock Training region-based object detectors with online hard example
  mining.
\newblock In {\em Proceedings of the IEEE conference on computer vision and
  pattern recognition}, pages 761--769, 2016.

\bibitem{sohn2016improved}
Kihyuk Sohn.
\newblock Improved deep metric learning with multi-class n-pair loss objective.
\newblock {\em Advances in neural information processing systems}, 29, 2016.

\bibitem{srinidhi2021survey}
Chetan~L Srinidhi, Ozan Ciga, and Anne~L Martel.
\newblock Deep neural network models for computational histopathology: A
  survey.
\newblock {\em Medical Image Analysis}, 67:101813, 2021.

\bibitem{suh2019him_2}
Yumin Suh, Bohyung Han, Wonsik Kim, and Kyoung~Mu Lee.
\newblock Stochastic class-based hard example mining for deep metric learning.
\newblock In {\em Proceedings of the IEEE/CVF Conference on Computer Vision and
  Pattern Recognition}, pages 7251--7259, 2019.

\bibitem{sun2019him_3}
Han Sun, Zhiyuan Chen, Shiyang Yan, and Lin Xu.
\newblock Mvp matching: A maximum-value perfect matching for mining hard
  samples, with application to person re-identification.
\newblock In {\em Proceedings of the IEEE/CVF International Conference on
  Computer Vision}, pages 6737--6747, 2019.

\bibitem{tan2022him_4}
Zichang Tan, Ajian Liu, Jun Wan, Hao Liu, Zhen Lei, Guodong Guo, and Stan~Z Li.
\newblock Cross-batch hard example mining with pseudo large batch for id vs.
  spot face recognition.
\newblock {\em IEEE Transactions on Image Processing}, 31:3224--3235, 2022.

\bibitem{tellez2019intro_pami_2}
David Tellez, Geert Litjens, Jeroen van~der Laak, and Francesco Ciompi.
\newblock Neural image compression for gigapixel histopathology image analysis.
\newblock {\em IEEE transactions on pattern analysis and machine intelligence},
  43(2):567--578, 2019.

\bibitem{tong2014multiple}
Tong Tong, Robin Wolz, Qinquan Gao, Ricardo Guerrero, Joseph~V Hajnal, Daniel
  Rueckert, Alzheimer’s Disease~Neuroimaging Initiative, et~al.
\newblock Multiple instance learning for classification of dementia in brain
  mri.
\newblock {\em Medical Image Analysis}, 18(5):808--818, 2014.

\bibitem{wang2018hsm}
Keze Wang, Xiaopeng Yan, Dongyu Zhang, Lei Zhang, and Liang Lin.
\newblock Towards human-machine cooperation: Self-supervised sample mining for
  object detection.
\newblock In {\em Proceedings of the IEEE Conference on Computer Vision and
  Pattern Recognition}, pages 1605--1613, 2018.

\bibitem{wang2019weakly}
Xi Wang, Hao Chen, Caixia Gan, Huangjing Lin, Qi Dou, Efstratios Tsougenis,
  Qitao Huang, Muyan Cai, and Pheng-Ann Heng.
\newblock Weakly supervised deep learning for whole slide lung cancer image
  analysis.
\newblock {\em IEEE transactions on cybernetics}, 50(9):3950--3962, 2019.

\bibitem{wang2018revisiting}
Xinggang Wang, Yongluan Yan, Peng Tang, Xiang Bai, and Wenyu Liu.
\newblock Revisiting multiple instance neural networks.
\newblock {\em Pattern Recognition}, 74:15--24, 2018.

\bibitem{wu2021combining}
Yunan Wu, Arne Schmidt, Enrique Hern{\'a}ndez-S{\'a}nchez, Rafael Molina, and
  Aggelos~K Katsaggelos.
\newblock Combining attention-based multiple instance learning and gaussian
  processes for ct hemorrhage detection.
\newblock In {\em MICCAI}, pages 582--591. Springer, 2021.

\bibitem{xu2019camel}
Gang Xu, Zhigang Song, Zhuo Sun, Calvin Ku, Zhe Yang, Cancheng Liu, Shuhao
  Wang, Jianpeng Ma, and Wei Xu.
\newblock Camel: A weakly supervised learning framework for histopathology
  image segmentation.
\newblock In {\em Proceedings of the IEEE/CVF International Conference on
  computer vision}, pages 10682--10691, 2019.

\bibitem{xu2019learning}
Lin Xu, Han Sun, and Yuai Liu.
\newblock Learning with batch-wise optimal transport loss for 3d shape
  recognition.
\newblock In {\em Proceedings of the IEEE/CVF Conference on Computer Vision and
  Pattern Recognition}, pages 3333--3342, 2019.

\bibitem{xu2014weakly}
Yan Xu, Jun-Yan Zhu, I Eric, Chao Chang, Maode Lai, and Zhuowen Tu.
\newblock Weakly supervised histopathology cancer image segmentation and
  classification.
\newblock {\em Medical Image Analysis}, 18(3):591--604, 2014.

\bibitem{zhang2022dtfd}
Hongrun Zhang, Yanda Meng, Yitian Zhao, Yihong Qiao, Xiaoyun Yang, Sarah~E
  Coupland, and Yalin Zheng.
\newblock Dtfd-mil: Double-tier feature distillation multiple instance learning
  for histopathology whole slide image classification.
\newblock In {\em Proceedings of the IEEE/CVF Conference on Computer Vision and
  Pattern Recognition}, pages 18802--18812, 2022.

\bibitem{Zhang_2022_BMVC}
Xiaoxian Zhang, Sheng Huang, Yi Zhang, Xiaohong Zhang, Mingchen Gao, and Liu
  Chen.
\newblock Dual space multiple instance representative learning for medical
  image classification.
\newblock In {\em 33rd British Machine Vision Conference 2022, {BMVC} 2022,
  London, UK, November 21-24, 2022}. {BMVA} Press, 2022.

\bibitem{zhao2022setmil}
Yu Zhao, Zhenyu Lin, Kai Sun, Yidan Zhang, Junzhou Huang, Liansheng Wang, and
  Jianhua Yao.
\newblock Setmil: spatial encoding transformer-based multiple instance learning
  for pathological image analysis.
\newblock In {\em Medical Image Computing and Computer Assisted
  Intervention--MICCAI 2022: 25th International Conference, Singapore,
  September 18--22, 2022, Proceedings, Part II}, pages 66--76. Springer, 2022.

\end{thebibliography}
}

\appendix

\section{Additional Visualization}
\begin{figure}[t]
\centering
    \includegraphics[width=\linewidth]{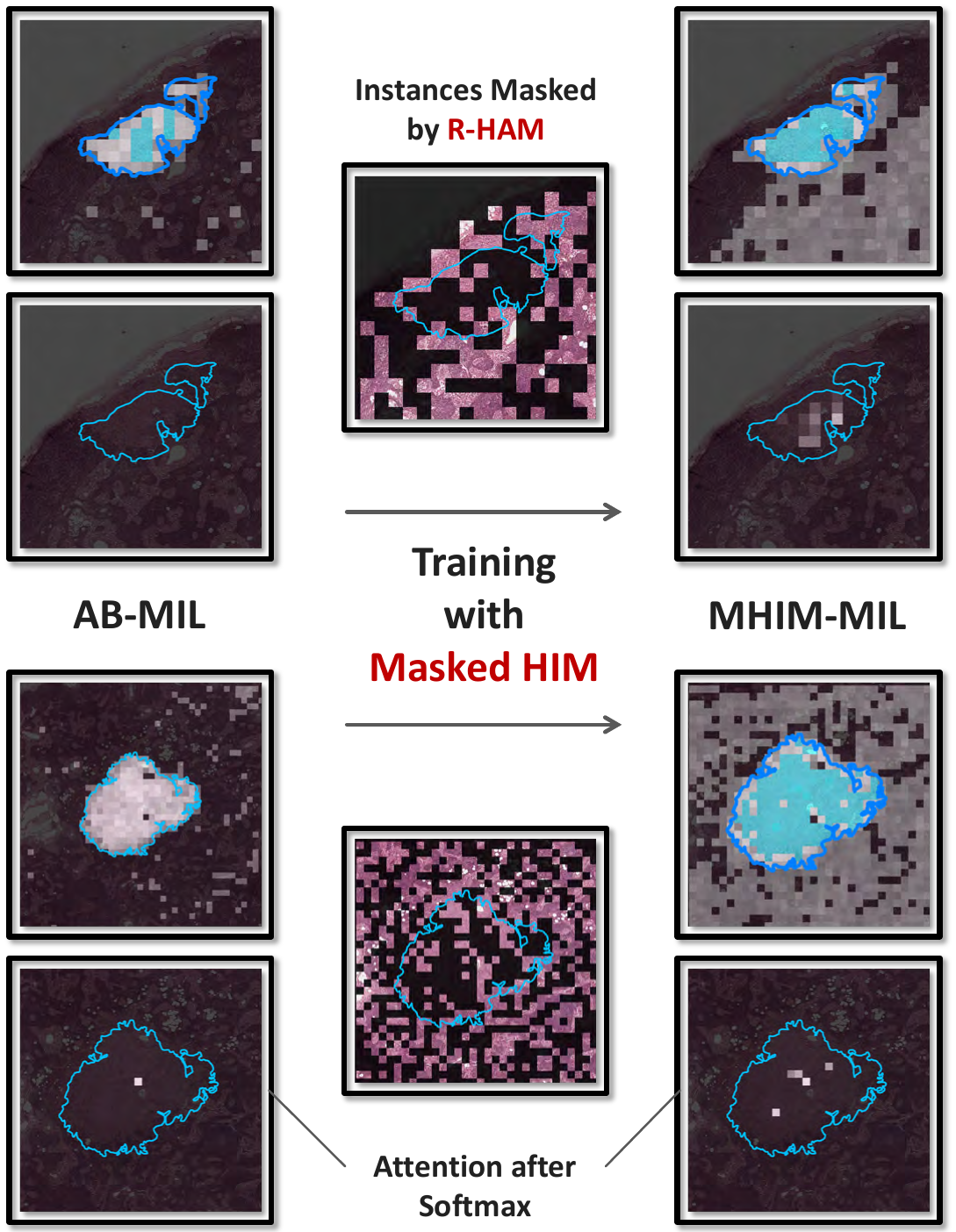}
    \caption{Comparison of patch visualization produced by AB-MIL~\cite{ilse2018attention} (baseline) and MHIM-MIL. The \textcolor{blue}{blue} lines outline the tumor regions. The brighter patch indicates higher attention scores. The \textcolor{cyan}{cyan} colors indicate high probabilities of being tumor for the corresponding locations. Ideally, the \textcolor{cyan}{cyan} patches should cover only the area within the \textcolor{blue}{blue} lines. In the middle column, the dark patches denote masked instances.}
    \label{fig:vis_spe}
\end{figure}

Here, we attempt to further analyze the impact of Masked Hard Instance Mining (Masked HIM) on WSI classification algorithms based on multiple instance learning. As shown in Figure~\ref{fig:vis_spe}, we visualize the masked instances (middle column), which we call the mined hard instances, to illustrate the relationship between the instance-level tumor prediction probability (\textcolor{cyan}{cyan} patch) and model attention (bright patch) before and after Masked HIM training. 

\begin{figure}[tb]
\centering
    \includegraphics[width=\linewidth]{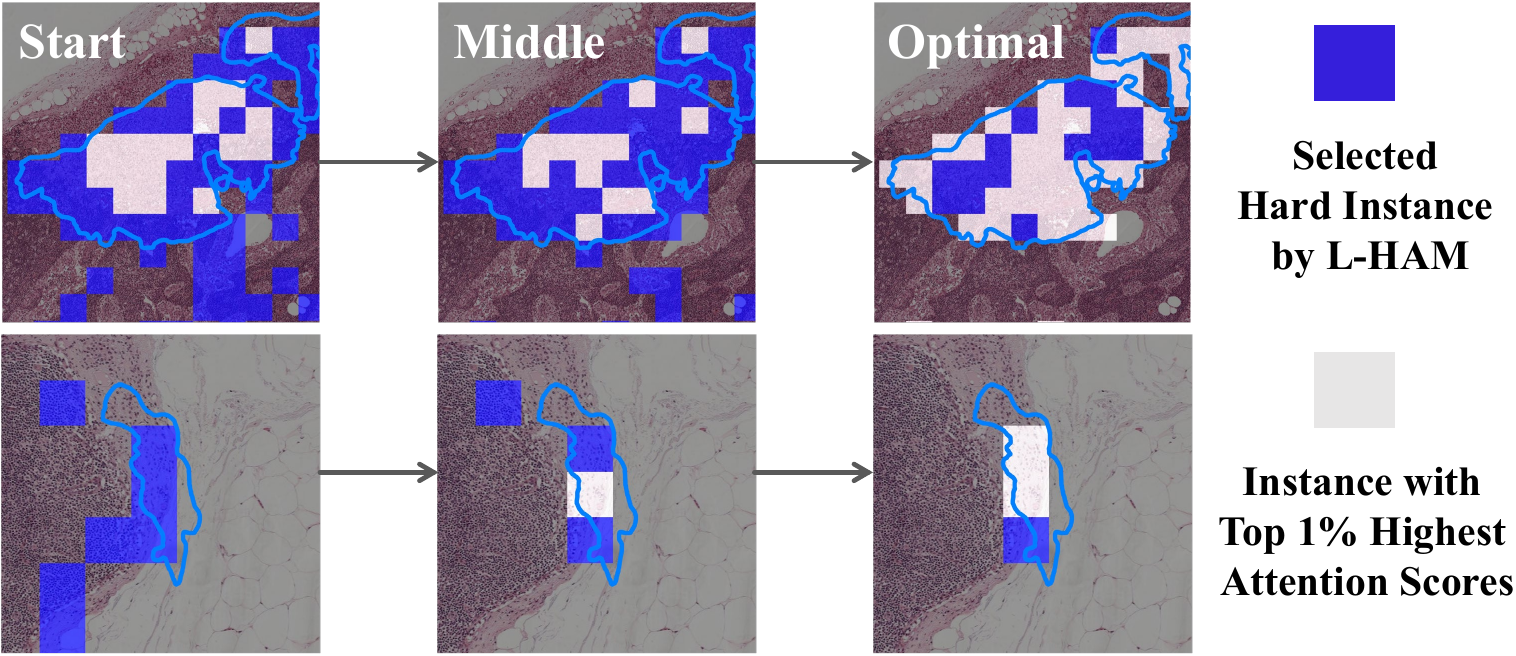}
    \caption{Patch visualization during iteration process.}
    \label{fig:vis_ds}
\end{figure}

First, thanks to the outstanding saliency patch mining ability of traditional attention-based MIL models, Masked HIM can effectively mask out the most salient regions to indirectly mine hard instances while using random masking to mitigate over-fitting problems. 
Moreover, as shown in the Figure~\ref{fig:vis_ds}, this discriminatory power improves gradually during the training.
To ensure that the instance sequence after masking still retains key instance information related to the slide category, we propose a randomization technique, which will be explained in detail in the following subsection.
Second, contrary to intuition, MIL models do not lose their discriminative power for key regions after masking out the most salient instances, due to the MHIM-MIL framework. Instead, they achieve a significant improvement.
Figure~\ref{fig:vis_spe} strongly proves that focusing only on salient instances during the training stage damages the discriminative power of MIL models, and verifies the huge help of hard instances for MIL model training. 
Moreover, we visualize the instance patch attention after softmax, which can be regarded as the contribution to the final bag embedding. 
We find that although traditional MIL models seem to pay attention to salient regions, they do not make reasonable use of this part of the information. 
They ignore most features and extremely focus on individual features in the feature aggregation process, damaging model discriminativeness. 
In contrast, MIL models trained with Masked HIM seem to put more attention on more “irrelevant regions”, but better utilize key region features to generate higher quality bag features and improve model performance.

\begin{figure*}[t]
    \begin{center}
    \includegraphics[width=\linewidth]{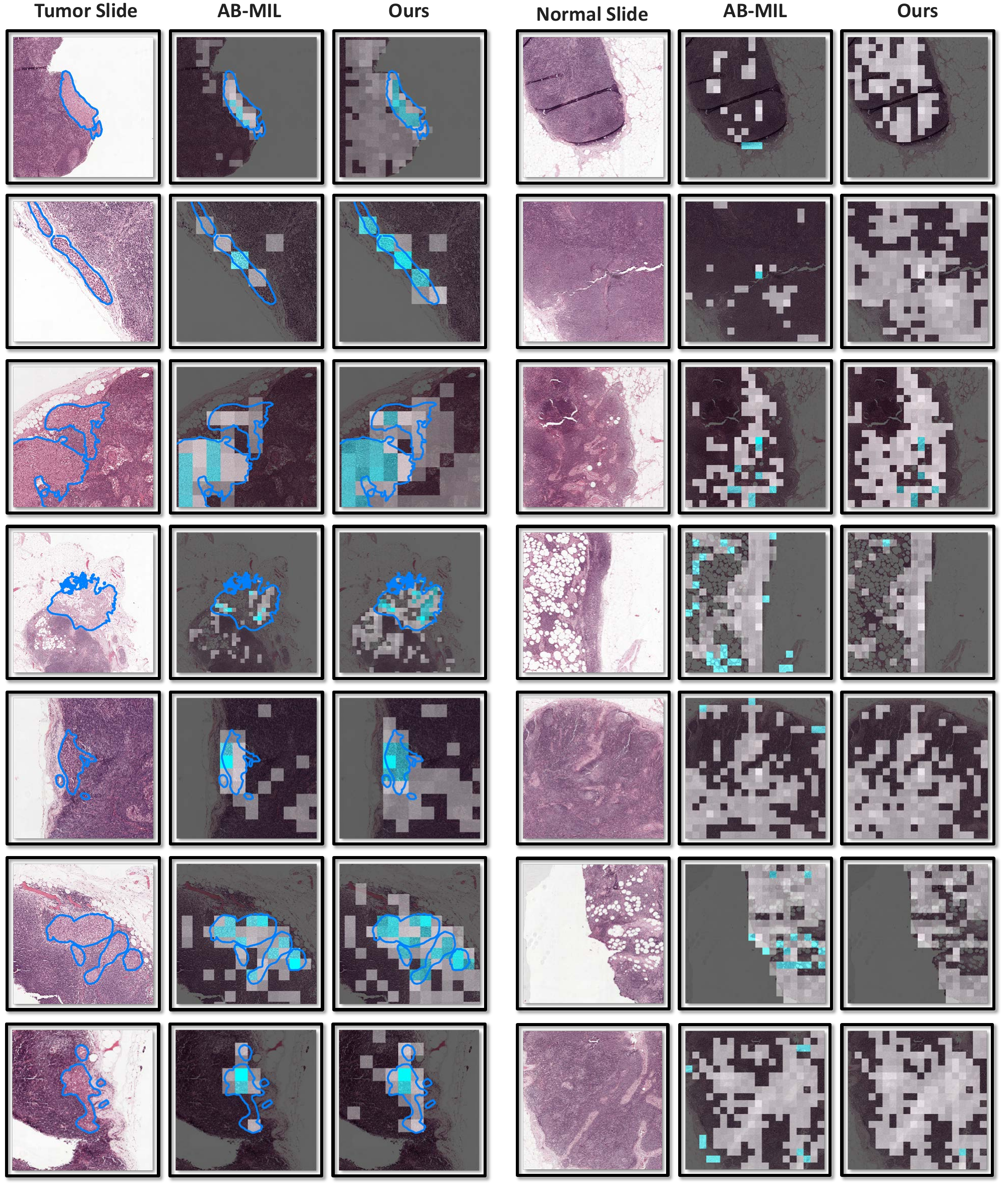}
    \end{center}
    \caption{More comparisons of patch visualization between AB-MIL (baseline) and MHIM-MIL. Best viewed in color.}
    \label{fig:big_vis}
\end{figure*}

Figure~\ref{fig:big_vis} shows more patch visualizations on the CAMELYON-16 dataset.

\begin{figure}[tb]
    \includegraphics[width=4.1cm]{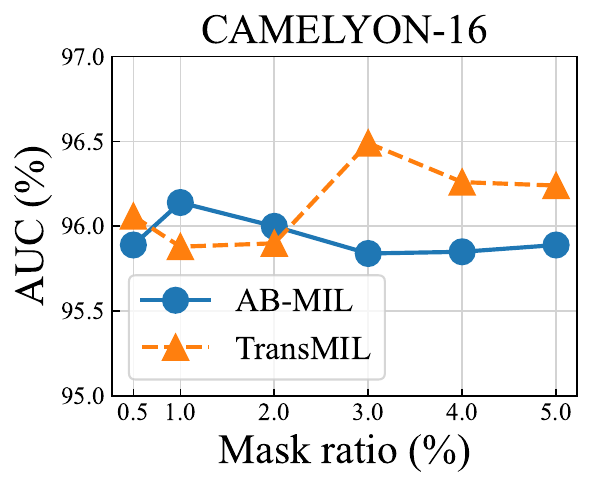}
    \includegraphics[width=4.1cm]{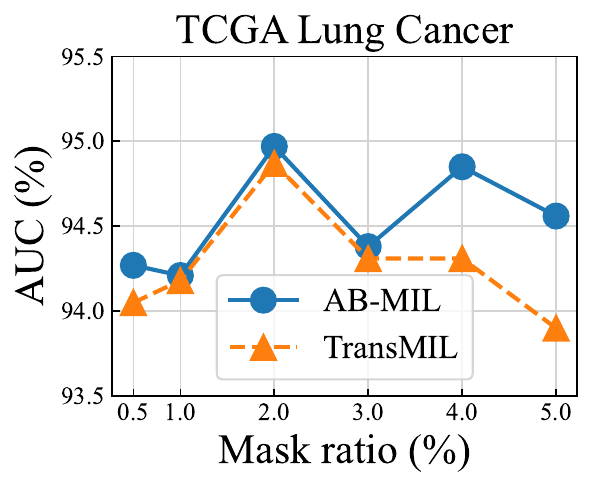}
    \caption{The performances of MHIM-MIL under different high attention mask ratio $\beta_{h}$.}
    \label{fig:mrh}
\end{figure}

\section{Additional Quantitative Experiments}
\subsection{More on Masked Hard Instance Mining}
\noindent\textbf{Discussion on Mask Ratio.}
We explored how various mask ratios affect MHIM-MIL training in this section. We fixed other ratios ($\beta_{r}$, $\beta_{l}$) and varied high attention mask ratio $\beta_{h}$ alone in Figure~\ref{fig:mrh}. We fixed $\beta_{h}$ and changed different $\beta_{r}$ and $\beta_{l}$ in Table~\ref{tab:mr}. Our findings are: 1) A low $\beta_{h}$ reduces the difficulty of mined instances, thereby diminishing the overall model performance. Moreover, the randomized trick ensures that the model training does not collapse even at high $\beta_{h}$. More details are provided in the following section. 2) Compared to AB-MIL~\cite{ilse2018attention}, TransMIL~\cite{shao2021transmil} has lower discriminative power for salient instances. This is why TransMIL needs a bigger $\beta_{h}$. 
3) MHIM-MIL training is less sensitive to $\beta_{r}$ and $\beta_{l}$ than to $\beta_{h}$. However, choosing an appropriate mask ratio is still crucial for optimal performance. Specifically, we observed that combining three strategies on the CAMELYON-16 dataset decreases classification performance. We attribute this to excessive instance masking losing important information on the CAMELYON-16 dataset.

\begin{table}[htb]
\centering
\begin{subtable}[b]{0.6\linewidth}
\begin{tabular}{ccc}
\toprule
   random ratio           & low ratio & AUC  \\ \midrule
 \textit{AB-MIL} && \\
60\%              &   20\%      & 94.57     \\
70\%             &   10\%    &  94.65       \\
\rowcolor{dino}\textbf{70\%}             &   \textbf{20\%}    &  \textbf{94.97}       \\
70\%             &   30\%    &  94.55       \\
80\%             &  20\%    & 94.49      \\\midrule
 \textit{TransMIL} &&   \\
50\%              & 20\%             & 94.60     \\
60\%             &  10\%       &   93.97       \\
\rowcolor{dino}\textbf{60\%}             &  \textbf{20\%}       &   \textbf{94.87}       \\
60\%             &  30\%       &   94.37       \\
70\%             &  20\%       &   94.60       \\

\bottomrule
\end{tabular}
\caption{TCGA Lung Cancer dataset}
\label{tab:tcga_mr}
\end{subtable}

\begin{subtable}[b]{0.6\linewidth}
\begin{tabular}{ccc}
\toprule
   random ratio           & low ratio & AUC  \\ \midrule
  \textit{AB-MIL} &&   \\ 
40\%              & 0\%             & 95.90     \\
\rowcolor{dino}50\%             &  0\%       &   96.14      \\
\textbf{50\%}              & \textbf{20\%}             & \textbf{95.92}     \\
60\%             & 0\%   & 96.13     \\ \midrule
 \textit{TransMIL} &&   \\
0\%              & 70\%             & 96.36     \\
\rowcolor{dino}\textbf{0\%}             &  \textbf{80\%}   & \textbf{96.49}      \\
20\%              & 80\%             & 96.33     \\
0\%             &   90\%    &  96.10    \\
\bottomrule
\end{tabular}
\caption{CAMELYON-16 dataset}
\label{tab:c16_mr}
\end{subtable}
\caption{Comparison of different random attention mask ratio $\beta_{r}$ and low attention mask ratio $\beta_{l}$ on both datasets.}
\label{tab:mr}
\end{table}

\noindent\textbf{Computational Cost.}
Here, we comprehensively discuss the impact of different MHIM strategies on the computational cost of model training. Table~\ref{tab:effi} shows the efficiency gains brought by large-scale low-attention masking and random-attention masking. This is especially significant for TransMIL~\cite{shao2021transmil}, a baseline with both spatial and temporal complexity quadratic to the number of instances. Large-scale masking greatly reduces the input of the student model, thereby reducing memory and time consumption. Although the input of the teacher model is still full length, due to the application of momentum teacher, it hardly introduces extra training cost. 
In addition, we also find that mixing multiple strategies further reduces the number of instances but also introduces additional computation, which is more obvious on AB-MIL~\cite{ilse2018attention} baseline.
\begin{table}[tb]
\centering
\begin{tabular}{lccccc}
\toprule
   Model           & C16 & TCGA & Para. & Time & Mem.\\ \midrule
\textit{AB-MIL} & 94.00 & 93.17 & 657K & \textbf{4.0s} & 2.4G \\ 
HAM & 95.68 & 93.83 &657K & \textbf{4.0s} & 2.7G \\
R-HAM & \textbf{96.14} & \textbf{94.79} &657K & 4.3s & 2.3G \\
L-HAM & 95.81 & 94.33 & 657K &  4.2s & 2.3G \\
LR-HAM & 95.92 & 94.97 &657K &  4.4s & \textbf{2.2G} \\
\midrule
\textit{TransMIL} & 93.51 & 92.51 &2.67M & 13.1s & 10.6G \\
HAM & 95.90 & 94.54 & 2.67M & 15.9s & 10.3G \\
R-HAM & 95.88 & 94.60 & 2.67M & 10.3s & 5.5G \\
L-HAM & \textbf{96.49} & 94.67 & 2.67M & \textbf{10.1s} & 5.5G \\
LR-HAM & 96.33 & \textbf{94.87} & 2.67M & \textbf{10.1s} & \textbf{5.4G} \\
\bottomrule
\end{tabular}
\caption{Comparison of time and memory requirements of different masked hard instance mining strategies. We report the model size (Para.), the training time per epoch (Time), and the peak memory usage (Mem.) on the CAMELYON-16 dataset (C16).}
\vspace{-0.3cm}
\label{tab:effi}
\end{table}

\noindent\textbf{Mask Ratio Decay.}
The discriminative ability of the model improves and stabilizes as training goes on. We follow the learning rate decay idea and tune $\beta_{h}$ based on training progress to prevent a high initial ratio from hurting later training. We name this technique mask ratio decay and adopt a classic cosine decay function to regulate decay speed. Table~\ref{tab:mrd} demonstrates that this trick significantly boosts performance. We note that we apply the decay strategy only to $\beta_{h}$ while maintaining initial values for the other two ratios during training.
\begin{table}[tb]
\centering
    \begin{tabularx}{0.87\linewidth}{p{1.7cm}cccc}
\toprule
{\multirow{2}{*}{Strategy}}  & \multicolumn{2}{c}{CAMELYON-16} & \multicolumn{2}{c}{TCGA} \\ \cmidrule(lr){2-3} \cmidrule(lr){4-5}
                & AB.         & Trans.        & AB.     & Trans.     \\\midrule
$\beta_{h}\%$               &   96.04   &  96.07  &  94.34 & 94.56 \\
\rowcolor{dino} \textbf{$\beta_{h}\%\rightarrow 0\%$}                  &   \textbf{96.14}      &   \textbf{96.49}       &  \textbf{94.97}    &    \textbf{94.87}     \\
\bottomrule 
\end{tabularx}
\caption{Comparison results of applying high attention mask ratio decay.}
\label{tab:mrd}
\end{table}

\begin{figure}[t]
\centering
    \includegraphics[width=\linewidth]{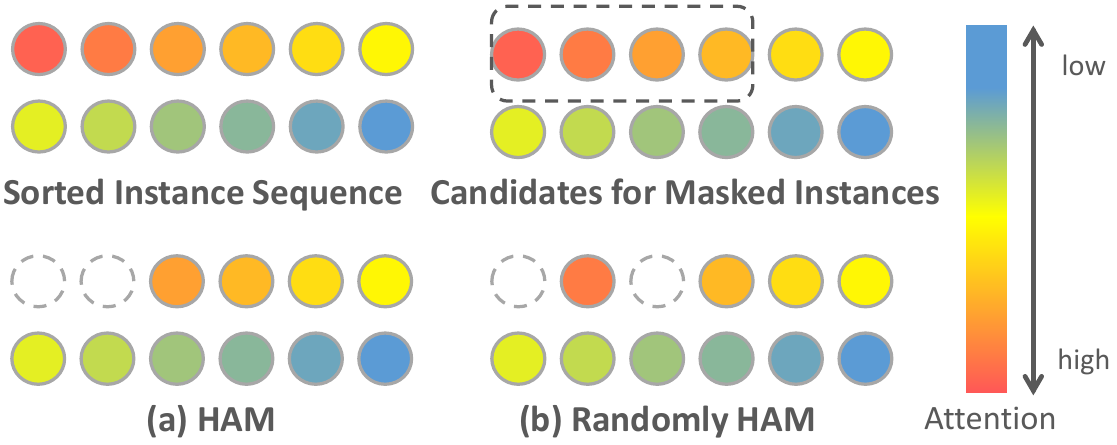}
    \caption{Illustration of Randomly High Attention Masking (Randomly HAM).}
    \label{fig:rham}
\end{figure}

\noindent\textbf{Randomly High Attention Masking.}
MHIM faces a major challenge: it may mask all key information and turn into ``error instance mining". We apply the Randomly High Attention Masking technique to address this issue and make sure that mined hard instances include key instance information for the slide category. Figure~\ref{fig:rham} illustrates our approach: we select instances with the highest $2\times \beta_{h}$\% attention scores as candidate states and randomly mask half of them to keep some key information. Table~\ref{tab:rham} demonstrates this technique suffers from low training difficulty in the TCGA dataset, where the tumor area ratio is high (typically over 40\%~\cite{shao2021transmil}), and impairs the discriminability of the model. On the other hand, this technique performs well on the CAMELYON-16 dataset, indicating that it can preserve key information in original instances.

\begin{table}[tb]
\centering
    \begin{tabularx}{1\linewidth}{lcccc}
\toprule
{\multirow{2}{*}{Strategy}}  & \multicolumn{2}{c}{CAMELYON-16} & \multicolumn{2}{c}{TCGA} \\ \cmidrule(lr){2-3} \cmidrule(lr){4-5}
                & AB.         & Trans.        & AB.     & Trans.     \\\midrule
w/o Ran. HAM                  &   95.71       &   96.37         & \textbf{94.97}     & \textbf{94.87}        \\
\rowcolor{dino}w/ Ran. HAM             & \textbf{96.14}        &\textbf{96.49}      & 94.52  & 94.17 \\

\bottomrule 
\end{tabularx}
\caption{Comparison results of applying randomly high attention masking (Ran. HAM).}
\label{tab:rham}
\end{table}

\subsection{Initialization of Student Network}
\begin{figure}[t]
\centering
    \includegraphics[width=0.9\linewidth]{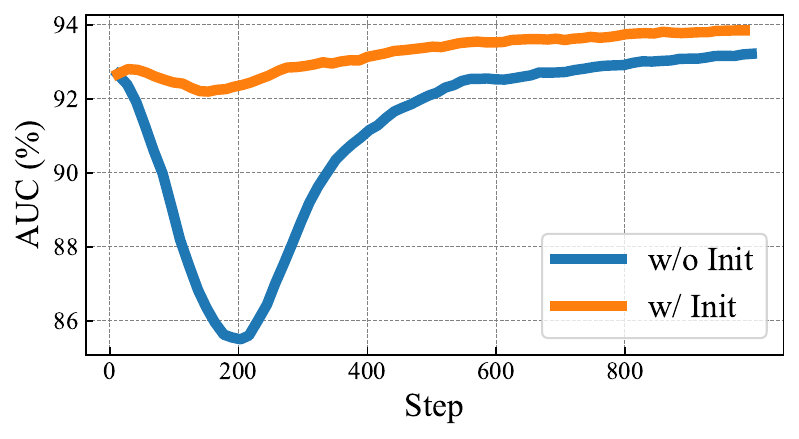}
    \caption{Performance comparison of teacher models under initialized or uninitialized student networks.}
    \label{fig:init_stu}
\end{figure}

MIL models typically employ a fully connected layer to project original 1024-dimensional instance features into 512 dimensions as final instance representation. 
In MHIM-MIL, we initialize the fully connected layer of the student network with pre-trained parameters to reduce collapse risk from the Siamese structure. 
\cite{caron2021dino} elaborates on more details about collapse risk. 
Figure~\ref{fig:init_stu} illustrates how this initialization affects teacher model performance. An uninitialized student model has slow initial training which drags down teacher model performance and harms the iterative optimization of the framework. The upper part of Table~\ref{tab:init_stu} displays a large margin in final student model performance with and without this initialization. Moreover, we applied the same initialization to mainstream MIL models to investigate if this initialization boosts performance by aiding Siamese structure optimization. 
The upper part of Table~\ref{tab:init_stu} reveals that this initialization does not noticeably enhance the performance of existing mainstream MIL models and sometimes lowers it. 
Our experiments confirm that initializing the first fully connected layer of student facilitate the iterative optimization of the MHIM-MIL framework instead of being a universal trick for increasing MIL model performance.
\begin{table}[htb]
\centering
\begin{tabular}{lcc}
\toprule
   Model           & CAMELYON-16 & TCGA  \\ \midrule
AB-MIL w/ init              & 93.98 (\textcolor{blue}{-0.02})            &  92.75 (\textcolor{blue}{-0.42})    \\

\rowcolor{dino}MHIM-MIL w/ init             & \textbf{96.14 (\textcolor{red}{+0.63})}       & \textbf{94.97 (\textcolor{red}{+0.49})}     \\\midrule

TransMIL w/ init             & 94.22 (\textcolor{red}{+0.71})       & 93.36 (\textcolor{red}{+0.85})     \\

\rowcolor{dino}MHIM-MIL w/ init             & \textbf{96.49 (\textcolor{red}{+0.90})}       & \textbf{94.87 (\textcolor{red}{+0.95})}     \\
\midrule
\textit{w/ init} &&\\
CLAM-SB             &  94.53 (\textcolor{blue}{-0.12})       &  93.43 (\textcolor{blue}{-0.24})      \\
DSMIL             &  94.96 (\textcolor{red}{+0.39})       &  93.93 (\textcolor{red}{+0.22})     \\
DTFD-MIL             & 95.23 (\textcolor{red}{+0.08})       & 93.80 (\textcolor{blue}{-0.03})     \\

\bottomrule
\end{tabular}
\caption{Comparison results of different initialized MIL models.}
\label{tab:init_stu}
\end{table}

\begin{figure}[t]
\centering
    \includegraphics[width=0.8\linewidth]{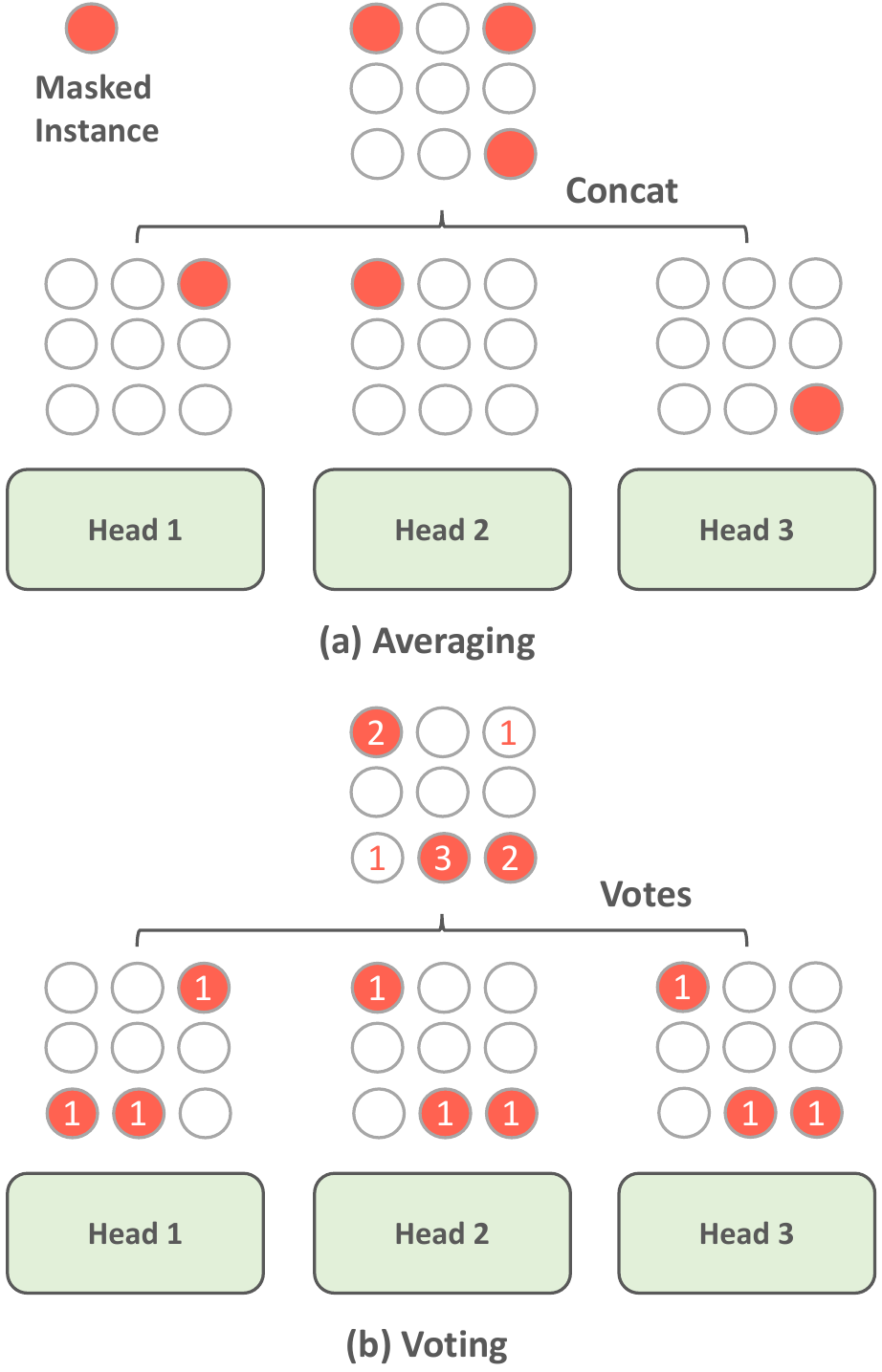}
    \caption{Illustration of averaging and voting multi-head attention fusion strategy.}
    \label{fig:vote}
\end{figure}

\subsection{Transformer Attention}

Transformer typically consists of a multi-layer multi-head structure where each head within each layer generates independent attention scores. Thus, extracting the most effective attention score among them is very challenging. 
In particular, the baseline model TransMIL~\cite{shao2021transmil} comprises two layers with eight heads per layer. We separately examined the effect of attention scores from different layers and various multi-head fusion strategies. 
The upper part of Table~\ref{tab:trans_attn} demonstrates the advantage of attention scores from the first layer over those from the final layer. 
We attribute this to the first layer producing more accurate attention scores for identifying hard instances. This is because the multi-head self-attention (MSA) operation modifies original features which causes a large deviation between hard instances mined by the last layer and the actual situation, while only the input of the first layer is the original instance features.

\begin{figure}[t]
\centering
    \includegraphics[width=0.8\linewidth]{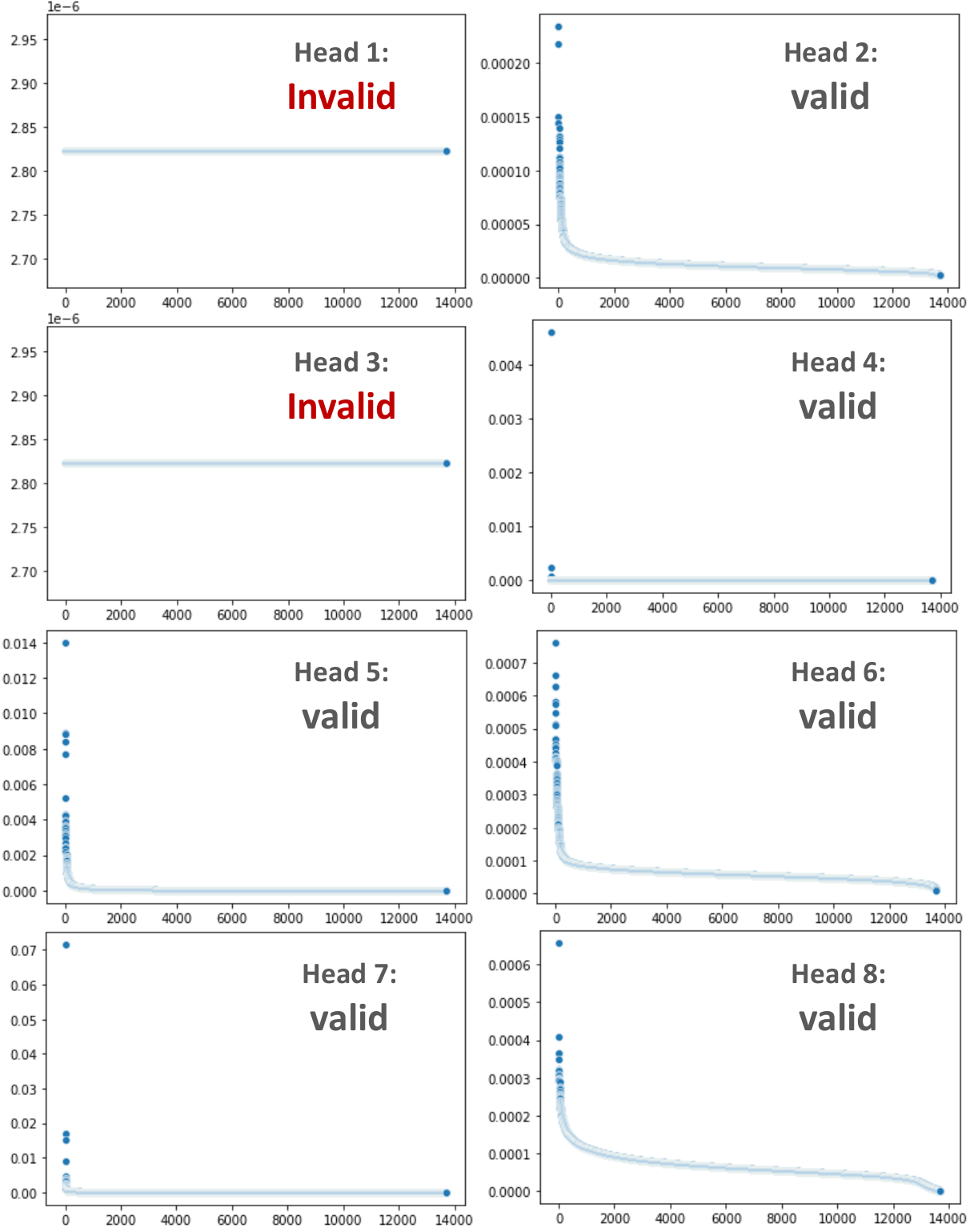}
    \caption{Attention visualization of different heads in TransMIL first layer.}
    \label{fig:vote_2}
\end{figure}

Additionally, prior work~\cite{he2022transfg} equalizes the contribution of each head and distributes the total mask count among different heads, which is called ``averaging". However, this strategy fails to prevent the effect of the invalid heads on MHIM. 
As shown in Figure~\ref{fig:vote_2}, some heads of TransMIL lack discrimination ability for instances and produce identical attention scores which we term as invalid heads. Invalid heads dilute localization accuracy for hard instances under averaging strategy and impair the training of MHIM-MIL. To mitigate this issue, we suggest a voting strategy that employs majority rule to eliminate noise from invalid heads, as shown in Figure~\ref{fig:vote}. The lower part of Table~\ref{tab:trans_attn} proves the effectiveness of this strategy.

\begin{table}[htb]
\centering
\begin{tabular}{lcc}
\toprule
   case           & CAMELYON-16 & TCGA  \\ \midrule
\rowcolor{dino}\textbf{first}              & \textbf{96.49}        &  \textbf{94.87}     \\
last             & 95.58 (\textcolor{blue}{-0.91})       & 93.90 (\textcolor{blue}{-0.97})    \\\midrule

averaging             &  96.38 (\textcolor{blue}{-0.11})       & 94.40 (\textcolor{blue}{-0.47})    \\
\rowcolor{dino}\textbf{voting}             & \textbf{96.49}       &  \textbf{94.87} \\

\bottomrule
\end{tabular}
\caption{Comparison results of variants of TransMIL attention.}
\label{tab:trans_attn}
\end{table}

\subsection{Discussion on Hyperparameter}
\begin{figure}[tb]
    \includegraphics[width=4.1cm]{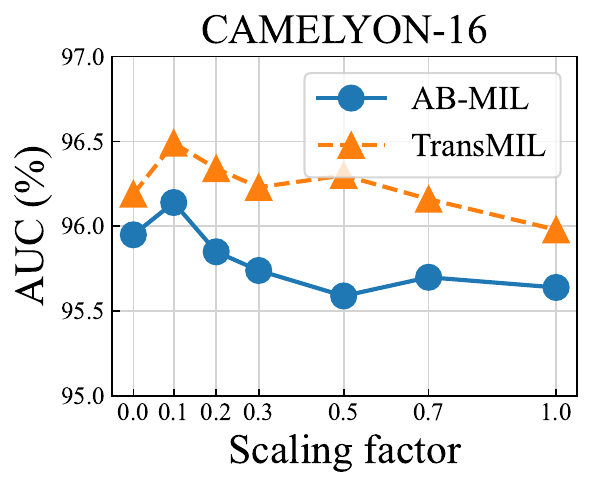}
    \includegraphics[width=4.1cm]{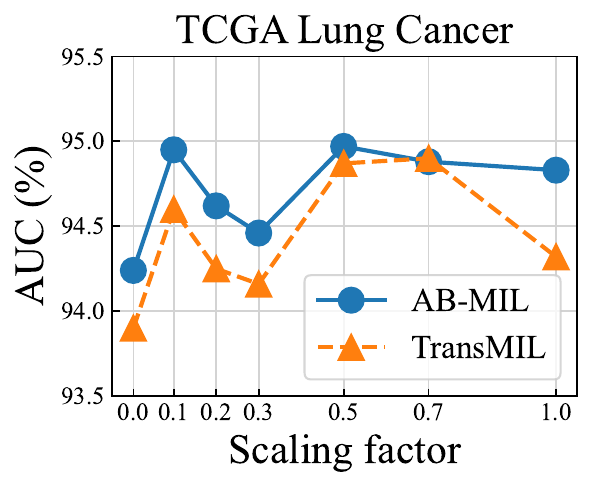}
    \caption{The performances of MHIM-MIL under different loss scaling factors $\alpha$.}
    \label{fig:loss}
\end{figure}

Here, we provide a systematic discussion of an important hyperparameter $\alpha$ in our framework. 
It balances the impact of self-supervised and fully supervised information during model training. Figure~\ref{fig:loss} demonstrates that $\alpha$ affects the training of both models consistently, with values that are either too high or too low resulting in biased training. Particularly, when $\alpha$ is too high, it impairs the positive effect of slide labels on model learning. This effect is more pronounced on the CAMELYON-16 dataset, as the model frequently misclassifies some challenging slides, requiring supervision from slide labels.

\section{Data Pre-processing}
Following prior works~\cite{clam,shao2021transmil,zhang2022dtfd}, we crop each WSI into a series of non-overlapping patches of size $256 \times 256$ at 20X magnification and discard the background region, including holes, as in CLAM~\cite{clam}. After pre-processing, we obtain a total of 3.6M patches from the CAMELYON-16 dataset, with an average of about 9000 patches per bag, and 10.8M patches from the TCGA Lung Cancer dataset, with an average of about 10300 patches per bag.



\section{Implementation Details}

Following~\cite{clam,shao2021transmil,zhang2022dtfd}, we use the ResNet-50 model~\cite{he2016deep} pretrained with ImageNet~\cite{deng2009imagenet} as the backbone network to extract an initial feature vector from each patch, which has a dimension of 1024. The last convolutional module of the ResNet-50 is removed, and a global average pooling is applied to the final feature maps to generate the initial feature vector. The initial feature vector is then reduced to a 512-dimensional feature vector by one fully-connected layer. 
The momentum rate of EMA is 0.9999 and the temperature of consistency loss is 0.1.
An Adam optimizer~\cite{kingma2014adam} with learning rate of $2\times 10^{-4}$ and weight decay of $1\times 10^{-5}$ is used for the model training. The Cosine strategy is adopted to adjust the learning rate.
All the models are trained for 200 epochs with an early-stopping strategy.
The patience of CAMELYON-16 and TCGA Lung Cancer 
are 30 and 20, respectively.
We do not use any trick to improve the model performance, such as gradient cropping or gradient accumulation.
The batch size is set to 1. All the experiments were conducted with an NVIDIA RTX3090 GPU.

\section{Pseudocode}
We present the PyTorch-style pseudocode for the training scheme of MHIM-MIL in Algorithm~\ref{algo:train}.

\section{Limitation}
In this paper, we propose a Masked Hard Instance Mining MIL framework to indirectly mine hard instances in the absence of instance supervision information. Although this strategy can effectively alleviate the over-reliance problem of traditional MIL models on salient instances, it is also challenging to accurately assess the difficulty level of instances and mine the most helpful hard instances for training. Compared with traditional hard sample mining strategies based on supervision information, this sub-optimal and rough strategy affects the convergence speed and discriminability of the model. In future work, we will focus on how to accurately evaluate instance difficulty level in the absence of complete supervision and use the most beneficial instances to facilitate model training.

\section{Code and Data Availability}
The source code of our project will be uploaded at~\href{https://github.com/DearCaat/MHIM-MIL}{https://github.com/DearCaat/MHIM-MIL}.

CAMELYON-16 dataset can be found at~\href{https://camelyon16.grand-challenge.org}{https://camelyon16.grand-challenge.org}.

TCGA Lung Cancer dataset can be found at~\href{https://portal.gdc.cancer.gov}{https://portal.gdc.cancer.gov}.

The script of slide pre-processing and patching can be found at~\href{https://github.com/mahmoodlab/CLAM}{https://github.com/mahmoodlab/CLAM}.


\setlength{\algomargin}{0em}
\SetAlFnt{\small}
\SetAlCapNameFnt{\small}

\begin{algorithm*}[t]
  \setstretch{0.8}
  \PyComment{f\underline{~}t, f\underline{~}s: teacher and student networks} \\
  \PyComment{f\underline{~}p: the pretrained network} \\
  \PyComment{mrh: high attention mask ratio} \\
  \PyComment{mrl: low attention mask ratio} \\
  \PyComment{mrr: random attention mask ratio} \\
  \PyComment{m: momentum rates} \\
  \PyComment{tp: temperatures} \\
  \PyComment{a: consistency loss scaling factor} \\
  ~\\
  \PyComment{initialize} \\
  \PyCode{f\underline{~}t.params = f\underline{~}p.params}\\
  \PyCode{f\underline{~}s.proj\underline{~}head.params = f\underline{~}p.proj\underline{~}head.params}\\
  ~\\
  \PyComment{teacher network not introduces any parameter} \\
  \PyCode{f\underline{~}t = f\underline{~}t.eval()}\\
  ~\\
  \PyCode{def mask\underline{~}fn(attn,mask\underline{~}ratio,largest):}\\
  \Indp
    \PyComment{sort attention score and get the topk index}\\
    \PyCode{attn = sort(attn)}\\
    \PyCode{topk\underline{~}ids = topk(attn,k=int(mask\underline{~}ratio*attn.length),largest=largest)}\\
    \PyComment{init vote matrix}\\
    \PyCode{vote = 0}\\
    \PyComment{voting and counting}\\
    \PyCode{vote[topk\underline{~}ids] = 1}\\
    \PyCode{vote = sum(vote)}\\
    \PyComment{get mask index}\\
    \PyCode{mask\underline{~}ids = topk(vote,k=int(mask\underline{~}ratio*attn.length))}\\
     ~\\
    \PyCode{return mask\underline{~}ids}\\
  \Indm
  ~\\
  \PyCode{for x,y in loader:} 
  \PyComment{load a minibatch x,y with N slides}\\
  \Indp   
    \PyComment{get attention scores from teacher}\\
    \PyCode{\underline{~},bag\underline{~}feats\underline{~}t,attn\underline{~}t = f\underline{~}t.forward(x)}\\
    \PyComment{stop gradient of teacher network}\\
    \PyCode{bag\underline{~}feats\underline{~}t = bag\underline{~}feats\underline{~}t.detach()}\\
    ~\\
    \PyComment{get masked instance index}\\
    \PyComment{High Attention Masking}\\
    \PyCode{mask\underline{~}h = mask\underline{~}fn(attn\underline{~}t,mrh,True)}\\
    \PyComment{Low Attention Masking}\\
    \PyCode{mask\underline{~}l = mask\underline{~}fn(attn\underline{~}t,mrl,False)}\\
    \PyComment{Random Attention Masking}\\
    \PyCode{mask\underline{~}r = random\underline{~}select(attn\underline{~}t,mrr)}\\
    \PyComment{Combine all index}\\
    \PyCode{mask\underline{~}all = mask\underline{~}h \& mask\underline{~}l \& mask\underline{~}r}\\
    ~\\
    \PyComment{masked hard instance mining}\\
    \PyCode{x\underline{~}hard = masking(x,mask\underline{~}all)}\\
    ~\\
    \PyCode{logits\underline{~}s,bag\underline{~}feats\underline{~}s,\underline{~} = f\underline{~}s.forward(x\underline{~}hard)}\\
    ~\\
    \PyComment{consistency loss}\\
    \PyCode{loss\underline{~}con = -softmax(bag\underline{~}feats\underline{~}t / tp) * log\underline{~}softmax(bag\underline{~}feats\underline{~}s)}\\
    \PyComment{label prediction loss}\\
    \PyCode{loss\underline{~}cls = CrossEntropy(logits\underline{~}s,y)}\\
    \PyCode{loss\underline{~}all = loss\underline{~}cls + a*loss\underline{~}con}\\
    ~\\
    \PyComment{Adam update: student network}\\
    \PyCode{loss\underline{~}all.backward()}\\
    \PyCode{update(f\underline{~}s.params)}\\
    ~\\
    \PyComment{EMA update: teacher network}\\
    \PyCode{f\underline{~}t.params = m*f\underline{~}t.params+(1-m)*f\underline{~}s.params}\\
    ~\\
    \PyComment{high attention mask ratio decay}\\
    \PyCode{CosineDecay(mrh)}\\
  \Indm 

    \caption{PyTorch-style pseudocode for MHIM-MIL training scheme}
    \label{algo:train}
\end{algorithm*}

\end{document}